\PassOptionsToPackage{dvipsnames,svgnames,x11names}{xcolor}
\documentclass[12pt]{article}

\usepackage{amsmath,amssymb,mathtools}
\usepackage[T1]{fontenc}
\usepackage[utf8]{inputenc}
\usepackage{lmodern}
\usepackage{microtype}
\usepackage{graphicx}
\usepackage{caption}
\usepackage{subcaption}
\usepackage{float}
\usepackage{enumitem}
\usepackage{xcolor}
\usepackage[normalem]{ulem}
\usepackage{amsthm}
\usepackage{algorithm}
\usepackage{algpseudocode}
\usepackage{tikz}
\usepackage{array}
\usepackage{booktabs}
\usetikzlibrary{calc,arrows.meta,positioning}
\usepackage{pgfplots}
\pgfplotsset{compat=1.18}

\usepackage[]{natbib}
\usepackage{bookmark}
\hypersetup{
  pdftitle={ReCIRC: Rectified Conformal Risk Control},
  colorlinks=true,
  linkcolor={blue},
  citecolor={Blue},
  urlcolor={Blue}
}

\renewcommand{\harvardurl}[1]{\url{#1}}

\theoremstyle{plain}
\newtheorem{theorem}{Theorem}[section]
\newtheorem{proposition}[theorem]{Proposition}
\newtheorem{corollary}[theorem]{Corollary}
\theoremstyle{definition}

\newtheorem{assumption}[theorem]{Assumption}
\theoremstyle{remark}
\newtheorem{remark}[theorem]{Remark}

\usepgfplotslibrary{groupplots}
\newcommand{\E}{\mathbb{E}}
\newcommand{\Prob}{\mathbb{P}}
\newcommand{\cD}{\mathcal{D}}
\newcommand{\cC}{\mathcal{C}}
\newcommand{\cX}{\mathcal{X}}
\newcommand{\cY}{\mathcal{Y}}
\newcommand{\loss}{\ell}
\newcommand{\Lam}{\Lambda}
\newcommand{\Rhat}{\widehat R}
\newcommand{\ind}{\mathbf{1}}
\newcommand{\recirc}{\texttt{ReCIRC}}

\newcommand{\anon}{1}

\begin{document}

\def\spacingset#1{\renewcommand{\baselinestretch}%
{#1}\small\normalsize} \spacingset{1}

\if1\anon
{
  \title{\bf ReCIRC: Rectified Conformal Risk Control}
  \author{
    Bruno Marcondes e Resende\thanks{Gratefully acknowledges support from FAPESP (grant 2025/04853-5)} \\
    Helton Graziadei \\
    Thiago Rodrigo Ramos\thanks{Gratefully acknowledges support from FAPESP (grant 2026/21108-4)} \\
    Rafael Izbicki\thanks{Gratefully acknowledges support from CNPq (grants 305065/2023-8 and 403458/2025-0) and FAPESP (grants 2023/07068-1 and 2025/24620-5).}
    \\
    Department of Statistics, Federal University of S\~ao Carlos
  }
  \maketitle
} \fi

\if0\anon
{
  \bigskip
  \bigskip
  \bigskip
  \begin{center}
    {\LARGE\bf ReCIRC: Rectified Conformal Risk Control}
\end{center}
  \medskip
} \fi

\bigskip
\begin{abstract}
Many applications of black-box predictive models require controlling task-relevant error rates, such as missed lesion pixels in segmentation or missed labels in multilabel classification. Conformal risk control \citep[CRC;][]{angelopoulos2024conformal} gives distribution-free guarantees for such losses, but it calibrates a single threshold shared by all inputs. Because conditional risk varies with the input,  this marginal guarantee often overprotects easy cases and underprotects hard ones.  We propose \recirc{} (Rectified Conformal Risk Control), which inverts each input's estimated local risk curve to reparameterize the calibrated threshold as a risk budget $a$ representing a common target conditional risk, and then applies CRC unchanged to the resulting family. \recirc{} retains CRC's finite-sample marginal guarantee regardless of the accuracy of the estimated curves, while  accurate curves yield approximate conditional risk control and, under additional conditions, asymptotically exact conditional risk  control; they also support a risk-calibration diagnostic. Across three synthetic and five real-data settings spanning segmentation, multilabel and multiclass classification, and regression, \recirc{} attained the lowest average worst-group risk and mean positive group excess in every setting, while maintaining marginal risk close to the target, whereas changes in prediction size were application-dependent.
\end{abstract}

\vfill

\newpage
\spacingset{1.8} 

\section{Introduction}
\label{sec:intro}

Black-box predictive models are now embedded in many consequential scientific and
clinical workflows \citep{wang2023scientific,topol2019high};
examples include segmentation networks that delineate polyps in colonoscopy
images, multilabel and hierarchical classifiers, and open-domain question-answering
systems \citep{fan2020pranet,lin2014microsoft,deng2009imagenet,kwiatkowski2019natural}.
In these applications,
point predictions alone are not enough. Practitioners need guarantees that
errors of specific, task-relevant kinds, such as missed lesion pixels, missed labels,
semantically distant predictions, occur at a controlled rate.

Conformal prediction provides such guarantees for the miscoverage of prediction
sets: wrapped around any predictive model, it produces sets that contain the
true response with a prescribed probability, without assumptions on the
data-generating distribution beyond exchangeability
\citep{vovk2005algorithmic,lei2018distribution,angelopoulos2023gentle}.
Conformal risk control \citep[CRC;][]{angelopoulos2024conformal} extends this
idea from miscoverage to arbitrary bounded losses that shrink as predictions
become more protective: given a family of decision rules
$\{f_\lambda:\lambda\in\Lam\}$ ordered so that larger $\lambda$ yields a
smaller loss $\loss(f_\lambda(x),y)$, CRC calibrates a data-driven threshold
$\widehat\lambda$ such that the expected loss of $f_{\widehat\lambda}$ on a
new covariate--response pair $(X,Y)$ is at most a target level $\alpha$. This covers, among others, the false-negative rate of
segmentation masks and label sets, graph-distance losses in hierarchical
classification, and F1-based losses in question answering.

Despite its generality, CRC calibrates a  single global parameter  shared
by every input. The core difficulty is that a raw threshold is not a risk
scale: the conditional risk ${R(\lambda\mid x)=\E\{\loss(f_\lambda(X),Y)\mid
X=x\}}$ typically varies with $x$, so the same $\lambda$ that yields nearly zero
conditional risk on easy inputs can leave hard inputs far above the target
level.  
Because the guarantee averages over the covariate distribution, it can
overprotect easy inputs while leaving difficult inputs above the target.

To address this limitation, we propose \emph{risk rectification}. The idea is
to keep the CRC calibration step untouched but change the parameter on which it
operates: instead of calibrating the raw threshold $\lambda$, we first convert
the original family into a new family indexed by a \emph{risk budget} $a$,
obtained by inverting each input's local risk curve. In the oracle version, the
same value of $a$ has the same interpretation at every input $x$: it is the
desired conditional expected loss. In practice, the local risk curves are
estimated, and ordinary CRC
is applied to the resulting reparameterized family to ensure distribution-free
guarantees.
We call the resulting framework \recirc{} (Rectified Conformal Risk Control).
The CRC
guarantee does not require the estimated risk curves to be correct: poor
estimates can reduce local adaptivity, but they cannot invalidate finite-sample
marginal risk control. Accurate estimates, in turn, make the calibrated budget
an approximately common conditional-risk scale, and we provide a graphical
goodness-of-fit diagnostic that targets exactly this property.

\begin{figure}[H]
    \centering
    \includegraphics[width=1\linewidth]{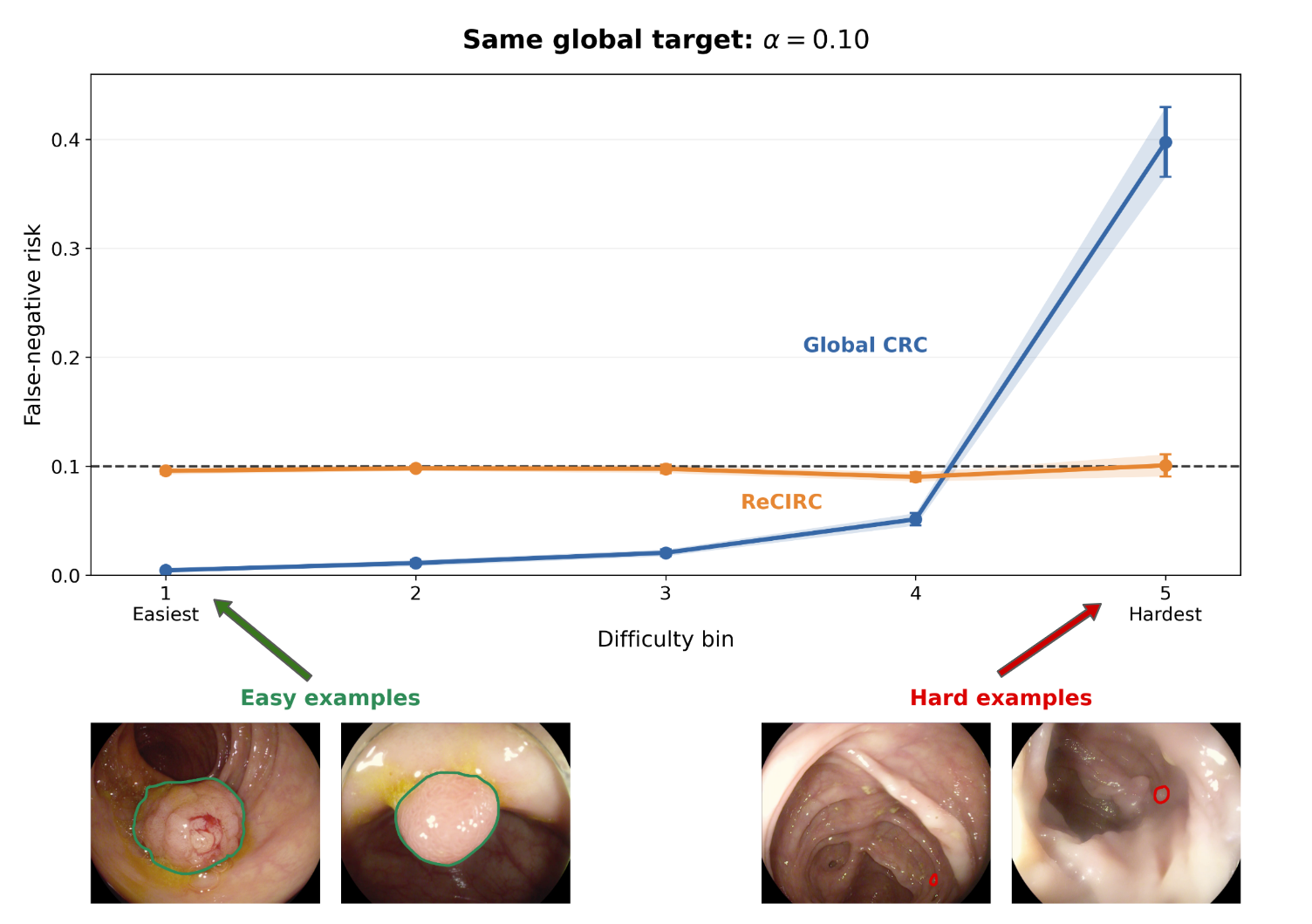}
    \caption{\textbf{Flattening the risk curve.} Conditional false-negative rate (FNR) by difficulty in the polyp-segmentation experiment (Section~\ref{sec:experiments}; target $\alpha=0.10$). Global CRC's FNR rises sharply on difficult images, whereas \recirc{} keeps it near the target. Curves show means and 95\% confidence intervals over 20 repetitions; images illustrate the easiest and hardest bins.}
    \label{fig:CRC_vs_ReCIRC_motivation}
\end{figure}

Figure~\ref{fig:CRC_vs_ReCIRC_motivation} illustrates this phenomenon in the
polyp-segmentation experiment. For each image, the false-negative rate (FNR) is
the fraction of ground-truth polyp pixels omitted from the predicted mask; the
conditional FNR shown in each difficulty bin is the average of this fraction
over the images in that bin. Global CRC concentrates these missed pixels among
difficult inputs, whereas \recirc{} keeps the conditional FNR closer to the target
across difficulty levels.

\textbf{Paper Outline.} 
Section~\ref{sec:background} reviews CRC; Sections~\ref{sec:method}
and~\ref{sec:theory} present \recirc{} and its guarantees; and
Section~\ref{sec:experiments} evaluates the method. Algorithms, proofs, and
worked examples appear in the appendices.

\subsection{Relation to Other Work}
\label{sec:related}

\textbf{Conformal methods for controlling prediction errors.}
Conformal prediction uses held-out calibration data to construct prediction
sets with finite-sample marginal coverage under exchangeability
\citep{vovk2005algorithmic,lei2018distribution,angelopoulos2023gentle}. CRC
extends this idea beyond set coverage: it controls the expected value of a
bounded loss that decreases as predictions become more protective \citep{angelopoulos2024conformal}.
Risk-controlling prediction sets and Learn-then-Test provide related guarantees
that hold with high probability over the calibration sample; the latter also
allows non-monotone losses and several tuning parameters
\citep{bates2021distribution,angelopoulos2021learn}. These guarantees are
marginal: they control error averaged across future inputs, not separately for
easy and difficult cases. \recirc{} addresses this heterogeneity by first indexing
each input's prediction by its estimated conditional risk and then applying a
standard calibration method. We use CRC, but the same \recirc{} family can be
calibrated by the other methods under their respective assumptions.

\textbf{Local and conditional conformal prediction.} Distribution-free conditional coverage cannot be
guaranteed in finite samples without admitting trivial prediction sets
\citep{lei2014distribution,vovk2012conditional,barber2021limits}. This
impossibility has motivated  a literature on approximate 
conditional validity. Mondrian and group-conditional constructions
\citep{vovk2012conditional} calibrate separately within user-specified
categories; multivalid calibration extends this to collections of possibly
overlapping groups \citep{jung2023batch}; localized calibration reweights the
calibration sample around the test point, with randomization restoring
robust guarantees \citep{guan2023localized,hore2025conformal}; data-driven
partitions replace the user's groups by ones learned from the data
\citep{cabezas2025regression}; and guarantees indexed by a function class
interpolate between the marginal and conditional extremes
\citep{gibbs2025conformal}. These approaches either coarsen the conditioning
target or modify how calibration itself is carried out. \recirc{} does
neither: it keeps ordinary calibration and instead learns a conditional
mean-loss scale, so that its finite-sample guarantee stays marginal and
distribution-free, while its pointwise and group-level statements are
inherited from the accuracy of the estimated risk curves.

\textbf{Score rectification and plug-in conditional calibration.} A recurring
idea in conformal prediction is to place heterogeneous inputs on a common scale
before calibrating a global cutoff. Examples include normalized scores
\citep{papadopoulos2011reliable}, conformalized quantile regression
\citep{romano2019conformalized}, and distribution-based transformations
\citep{chernozhukov2021distributional,izbicki2020flexible,izbicki2022cd,han2022split}.
Most directly, \citet[Eq.~14]{dheur2025unified} transform a conformity score
through its estimated conditional CDF, while EPICSCORE
\citep{cabezas2025epistemic} uses a posterior predictive CDF to incorporate
epistemic uncertainty. \citet{plassier2025rectifying} instead learn a
transformation from a conditional quantile at the target coverage level.
Section~\ref{sec:score-rectification} shows that \recirc{} recovers the
conditional-CDF construction under the miscoverage loss function. Our method extends this
principle to bounded losses that are monotone along ordered decision families:
it estimates the conditional mean-loss curve, inverts it to define a common
risk scale, and calibrates that scale with ordinary CRC. The same fitted curve
can be recalibrated at multiple risk targets, and its estimation error affects
the conditional-risk bounds but not finite-sample marginal validity.

\textbf{Adaptive conformal risk control.} AA-CRC is the closest method in aim:
for a fixed target $\alpha$, it learns an input-dependent threshold and controls
weighted risk over every non-negative weight in a chosen vector space of
functions, up to a regularization term
\citep{blot2024automatically,gibbs2025conformal}. \recirc{} instead estimates the
entire conditional mean-loss curve on independent data, uses it to define a
one-parameter local risk scale, and delegates final calibration to ordinary
CRC. Its pointwise and group-level conclusions are expressed directly through
risk-estimation error (Propositions~\ref{prop:approx-conditional}
and~\ref{prop:aggregate-conditional}), while
Corollary~\ref{cor:deployed-conditional} gives conditions under which the
deployed conditional risk converges to the target. Fair Risk Control targets
multi-group fairness constraints through multicalibration
\citep{zhang2024fair}. Localized adaptive risk control instead operates online,
updating a threshold function in a reproducing kernel Hilbert space
\citep{zecchin2024localized}; \recirc{} is an offline procedure whose final
predictor retains CRC's finite-sample marginal guarantee.

\textbf{Risk-based parameterizations for specific tasks.} Two task-specific
constructions use a similar reparameterization to the one we propose.
\citet{xu2023ordinal} study ordinal classification, where prediction sets are
contiguous ranges of classes. They index the nested family by a plug-in
estimate of the conditional risk, computed from the classifier's estimated
class probabilities, and then calibrate that index with CRC. \citet{luo2025conditional}
pursue the same goal in image segmentation, adapting masks through the
predicted pixel-probability mass, with a probability-calibrated variant and a
stratified variant that assigns group-specific thresholds; only the latter
requires a group structure. \recirc{} formulates risk rectification for general
ordered families and arbitrary bounded monotone losses.

\subsection{Novelty}
\label{sec:novelty}

Our contributions are:
\begin{itemize}
\item \textbf{A general risk-scale construction:} We formulate local
risk-curve inversion for bounded monotone losses along ordered decision
families, extending risk-based parameterizations beyond individual tasks.

\item \textbf{Conditional guarantees from estimation accuracy:} We relate
risk-curve estimation error to pointwise, most-input, and group-conditional
risk bounds relative to the budget
(Propositions~\ref{prop:approx-conditional} and~\ref{prop:aggregate-conditional}).
Under additional attainability and regularity conditions, we obtain a rate
for convergence of deployed conditional risk to the target
(Corollary~\ref{cor:deployed-conditional}).

\item \textbf{A risk-scale diagnostic:}  We compare realized losses with
advertised budgets, overall and within covariate groups, to assess the fitted
scale (Section~\ref{sec:gof}). The diagnostic is a by-product of calibration
and requires no data beyond the calibration set.
\end{itemize}

Distribution-free marginal validity follows by applying the existing CRC calibration argument
to the rectified family (Theorem~\ref{thm:marginal-validity};
\citealp{angelopoulos2024conformal}), regardless of risk-estimation accuracy.

\section{Background: Conformal Risk Control}
\label{sec:background}

\subsection{Setting and the CRC procedure}
\label{sec:crc}

Let $(X_1,Y_1),\ldots,(X_n,Y_n),(X_{n+1},Y_{n+1})$ be exchangeable random pairs
taking values in $\cX\times\cY$, where the first $n$ pairs form a calibration
set $\cC$ and $(X_{n+1},Y_{n+1})$ is a test point whose response is unobserved.
Consider a family of decision rules $\{f_\lambda:\lambda\in\Lam\}$ indexed by a
scalar $\lambda\in\Lam=[\lambda_{\min},\lambda_{\max}]$, together with a loss
$\loss(f_\lambda(x),y)\in[0,B]$. The family is assumed to be ordered so that
larger values of $\lambda$ are more protective:
\begin{equation}
  \lambda_1\le \lambda_2
  \quad\Longrightarrow\quad
  \loss(f_{\lambda_1}(x),y)\ge
  \loss(f_{\lambda_2}(x),y)
  \quad\text{for every }(x,y).
  \label{eq:monotone-family}
\end{equation}
The prototypical example takes $f_\lambda(x)$ to be a prediction set that grows
with $\lambda$ and $\loss$ to be a loss that shrinks as the set grows, such as
the false-negative proportion of a segmentation mask or a label set.
 For example, let $Y\subseteq\mathcal U$ denote the true positive pixels or
labels and let $f_\lambda(x)\subseteq\mathcal U$ be the selected set. The loss
\[
  \loss_{\rm FN}(f_\lambda(x),Y)
  :=\frac{|Y\setminus f_\lambda(x)|}{|Y|\vee1}
\]
is bounded by one and is non-increasing as $f_\lambda(x)$ expands.

Throughout,
the family $\{f_\lambda\}$ is treated as fixed: when it is built from a trained
base model, all expectations and risk curves below are interpreted
conditionally on the data used to fit it, which is assumed independent of the
calibration and test points.

Given a target risk level $\alpha$, CRC \citep{angelopoulos2024conformal} computes the calibration risk
$\widehat{\mathcal R}_{\mathrm{cal}}(\lambda)
=n^{-1}\sum_{j=1}^n \loss(f_\lambda(X_j),Y_j)$ and selects
\begin{equation*}
  \widehat\lambda
  =
  \inf\left\{\lambda\in\Lam:
  \frac{n}{n+1}\,\widehat{\mathcal R}_{\mathrm{cal}}(\lambda)
  +\frac{B}{n+1}
  \le \alpha
  \right\}.
\end{equation*}
Under exchangeability, monotonicity, and right-continuity of the loss in
$\lambda$, \citet{angelopoulos2024conformal} show that
\begin{equation}
  \E\{\loss(f_{\widehat\lambda}(X_{n+1}),Y_{n+1})\}
  \le \alpha,
  \label{eq:crc-guarantee}
\end{equation}
and that this bound is nearly tight, the expected loss being at least
$\alpha-2B/(n+1)$ under mild regularity conditions. When $\loss$ is the
miscoverage loss $\ind\{y\notin f_\lambda(x)\}$, \eqref{eq:crc-guarantee}
recovers the marginal coverage guarantee of split conformal prediction.

\subsection{The limitation: a global threshold is not a risk scale}
\label{sec:limitation}

\begin{figure}[t!]
\centering

\definecolor{crcblue}{HTML}{3667A6}
\definecolor{recircorange}{HTML}{E68632}
\definecolor{easygreen}{HTML}{2E8B57}
\definecolor{hardred}{HTML}{DB0000}
\begin{tikzpicture}
\begin{groupplot}[
    width=3.9cm,
    height=3.3cm,
    scale only axis,
    group style={group size=3 by 1, horizontal sep=1.45cm},
    xmin=0, xmax=1,
    ymin=0, ymax=1,
    axis lines=left,
    title style={font=\small},
    label style={font=\small},
    xlabel style={at={(axis description cs:0.5,-0.2)}, anchor=north},
    ticklabel style={font=\scriptsize},
    samples=160,
    domain=0:1,
]
\nextgroupplot[
    title={(a) Same $\lambda$, different risks},
    xlabel={$\lambda$},
    ylabel={$R(\lambda\mid x)$},
    xtick={0,0.45,1},
    xticklabels={$\lambda_{\min}$,$\lambda_0$,$\lambda_{\max}$},
    ytick={0,1},
    yticklabels={$0$,$B$},
]
\addplot[easygreen, thick] {0.8*(1-x)^3};
\addplot[hardred, thick, dashed] {0.9*(1-x)^0.6};
\addplot[gray, dotted, thick] coordinates {(0.45,0) (0.45,1)};
\addplot[easygreen, mark=*, only marks] coordinates {(0.45,0.1331)};
\addplot[hardred, mark=*, only marks] coordinates {(0.45,0.6287)};
\node[font=\scriptsize, easygreen, anchor=south west] at (axis cs:0.58,0.07) {easy};
\node[font=\scriptsize, hardred, anchor=south west] at (axis cs:0.74,0.44) {hard};
\nextgroupplot[
    title={(b) Same risk, different $\lambda$},
    xlabel={$\lambda$},
    ylabel={$R(\lambda\mid x)$},
    xtick={0.279,0.840},
    xticklabels={$R^{-1}(a\mid x_{\rm easy})$,$R^{-1}(a\mid x_{\rm hard})$},
    ytick={0,0.30,1},
    yticklabels={$0$,$a$,$B$},
]
\addplot[easygreen, thick] {0.8*(1-x)^3};
\addplot[hardred, thick, dashed] {0.9*(1-x)^0.6};
\addplot[gray, dotted, thick] coordinates {(0,0.30) (1,0.30)};
\addplot[easygreen, dotted, thick] coordinates {(0.279,0) (0.279,0.30)};
\addplot[hardred, dotted, thick] coordinates {(0.840,0) (0.840,0.30)};
\addplot[easygreen, mark=*, only marks] coordinates {(0.279,0.30)};
\addplot[hardred, mark=*, only marks] coordinates {(0.840,0.30)};
\node[font=\scriptsize, easygreen, anchor=south west] at (axis cs:0.58,0.07) {easy};
\node[font=\scriptsize, hardred, anchor=south west] at (axis cs:0.74,0.44) {hard};
\nextgroupplot[
    title={(c) Risk across inputs},
    xlabel={inputs, easy $\to$ hard},
    ylabel={conditional risk},
    xtick={0.333,0.878},
    xticklabels={$x_{\rm easy}$,$x_{\rm hard}$},
    ytick={0,0.30,1},
    yticklabels={$0$,$\alpha$,$B$},
    legend style={font=\scriptsize, draw=none, fill=none, at={(0.02,0.98)}, anchor=north west},
    legend cell align=left,
]
\addplot[crcblue, thick] {0.05+0.75*x^2};
\addlegendentry{global threshold $\lambda_0$}
\addplot[recircorange, thick] coordinates {(0,0.30) (1,0.30)};
\addlegendentry{rectified, $a=\alpha$}
\draw[easygreen, thick, dotted, -{Stealth[length=1.5mm]}] (axis cs:0.333,0.133) -- (axis cs:0.333,0.29);
\draw[hardred, thick, dotted, -{Stealth[length=1.5mm]}] (axis cs:0.878,0.628) -- (axis cs:0.878,0.31);
\addplot[easygreen, mark=*, only marks] coordinates {(0.333,0.133)};
\addplot[hardred, mark=*, only marks] coordinates {(0.878,0.628)};
\end{groupplot}
\end{tikzpicture}

\caption{\textbf{Risk rectification as a reparameterization.} (a) The same raw
threshold $\lambda_0$ gives different conditional risks at an easy and a hard
input. (b) A common risk budget $a$ instead gives each input its own threshold
$R^{-1}(a\mid x)$, obtained by inverting its local risk curve. (c) Conditional
risk across inputs, with the two inputs of (a) marked. A global threshold whose
average risk is $\alpha$ (here $\lambda_0$) leaves easy inputs below $\alpha$
and hard inputs above it; oracle rectification with $a=\alpha$ brings every
input to $\alpha$. In practice the curves are estimated and CRC calibrates $a$
(Section~\ref{sec:procedure}). The plots are normalized to $B=1$.}
\label{fig:risk-rectification}
\end{figure}
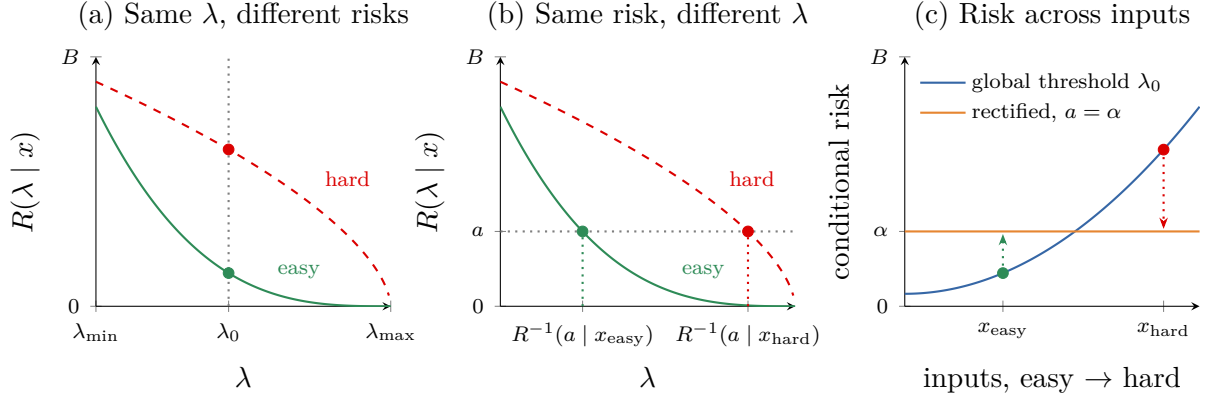

The guarantee in \eqref{eq:crc-guarantee} is marginal: it averages over the
covariate distribution. To see what it hides, define the \emph{local
conditional risk curve} at input $x$,
\begin{equation}
  R(\lambda\mid x)
  :=
  \E\{\loss(f_\lambda(X),Y)\mid X=x\},
  \label{eq:local-risk-curve}
\end{equation}
which is non-increasing in $\lambda$ by \eqref{eq:monotone-family}. CRC
controls only the average $\E\{R(\widehat\lambda\mid X)\}$, and
$R(\lambda\mid x)$ is typically far from constant in $x$: the same raw
threshold $\lambda$ intersects the risk curves of easy and hard inputs at very
different risk levels, as illustrated in
Figure~\ref{fig:risk-rectification}(a). A value of $\lambda$ that gives low
risk for one input may give high risk for another, so risk can fall below the target on easy inputs and exceed it on hard ones
(Figure~\ref{fig:risk-rectification}(c)). This is
the problem that risk rectification addresses.

\section{Our approach: \recirc{}}
\label{sec:method}

The key insight of \recirc{} is to reinterpret the parameter calibrated by CRC
as a local risk budget: instead of asking every input to share the same raw
threshold $\lambda$, we ask every input to share the same target conditional
risk $a$, and let the threshold adapt to the input. We first describe the
oracle construction, which assumes the local risk curves are known, and then
the practical procedure, which estimates them.

\subsection{Oracle risk rectification}
\label{sec:oracle}

Suppose for a moment that the local risk curves $R(\lambda\mid x)$ in
\eqref{eq:local-risk-curve} were known. For a desired risk budget $a\in[0,B]$,
define the risk-curve inverse
\begin{equation*}
  R^{-1}(a\mid x)
  :=
  \inf\{\lambda\in\Lam:R(\lambda\mid x)\le a\},
\end{equation*}
with the convention $\inf\emptyset=\lambda_{\max}$. Throughout,
$R^{-1}(a\mid x)$ denotes this generalized inverse of the local curve in the
threshold coordinate: its input is a risk level $a$, and its output is the raw
decision threshold needed at input $x$ to attain conditional risk at most $a$.
Figure~\ref{fig:risk-rectification}(b) shows this inversion: a horizontal cut
at risk level $a$ produces one threshold for an easy input and a larger, more
protective threshold for a hard input.

We define the \emph{oracle rectified family} by
\begin{equation}
  f_a^R(x)
  :=
  f_{R^{-1}(a\mid x)}(x),
  \qquad a\in[0,B].
  \label{eq:oracle-rectified}
\end{equation}
The family $\{f_a^R:a\in[0,B]\}$ is simply a reparameterization of the original
family $\{f_\lambda:\lambda\in\Lam\}$: each rectified rule outputs an element
of the original nested family, so the geometry of the predictions is inherited
from the original CRC construction (Appendix~\ref{app:examples} illustrates
this for several losses from the CRC literature). The difference is that the
new parameter $a$ has a direct risk interpretation. Assume that
$R(\cdot\mid x)$ is right-continuous, as holds whenever the loss is
right-continuous in $\lambda$ (the standard CRC requirement of
Section~\ref{sec:crc}). The construction then gives
\begin{equation}
  \E\{\loss(f_a^R(X),Y)\mid X=x\}
  =
  R(R^{-1}(a\mid x)\mid x)
  \le a,
  \label{eq:oracle-bound}
\end{equation}
so, in the oracle case, $a$ is a common upper bound on the conditional risk at
\emph{every} input $x$ (Figure~\ref{fig:risk-rectification}(c)). If the local curve is continuous and crosses level
$a$, then the achieved risk equals $a$; otherwise it may be conservative.
Thus risk rectification changes the global CRC knob from a raw threshold
$\lambda$, which is not comparable across inputs, into a common conditional-risk
upper bound and, where the curves cross the target, an exact risk scale.

\subsection{The \recirc{} procedure}
\label{sec:procedure}

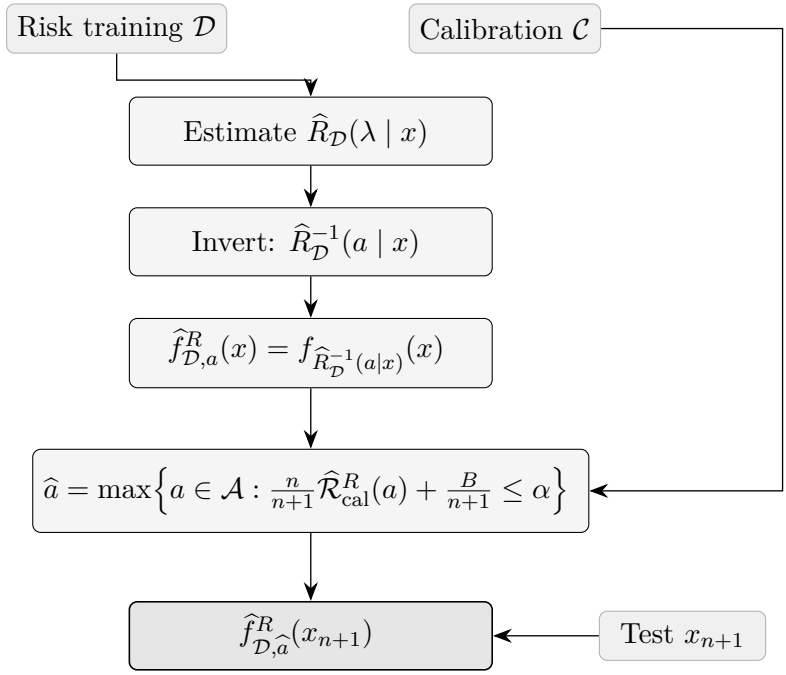
\begin{figure}[t!]
\centering
\begin{tikzpicture}[
    font=\small,
    >={Stealth[length=2.5mm]},
    every node/.style={align=center},
    block/.style={
        rectangle, rounded corners=3pt,
        draw=black, line width=0.4pt,
        fill=black!4,
        minimum width=4.8cm, minimum height=0.9cm,
        inner sep=4pt
    },
    data/.style={
        rectangle, rounded corners=3pt,
        draw=black!30, line width=0.4pt,
        fill=black!6,
        minimum width=2.2cm, minimum height=0.65cm
    },
    output/.style={
        rectangle, rounded corners=3pt,
        draw=black, line width=0.6pt,
        fill=black!10,
        minimum width=4.8cm, minimum height=0.9cm
    },
    arr/.style={->, draw=black, line width=0.5pt}
]
\node[data] (D) {Risk training $\mathcal{D}$};
\node[data, right=2.4cm of D] (C) {Calibration $\mathcal{C}$};
\node[block, below=0.9cm of $(D)!0.5!(C)$] (estimate) {
    Estimate $\widehat{R}_{\mathcal{D}}(\lambda \mid x)$
};
\node[block, below=0.55cm of estimate] (invert) {
    Invert: $\widehat{R}_{\mathcal{D}}^{-1}(a \mid x)$
};
\node[block, below=0.55cm of invert] (family) {
    $\widehat{f}^{R}_{\mathcal{D},a}(x) = f_{\widehat{R}_{\mathcal{D}}^{-1}(a\mid x)}(x)$
};
\node[block, below=0.8cm of family, minimum height=1.1cm] (crc) {
    $\widehat{a} = \max\!\left\{ a\in\mathcal{A} : \tfrac{n}{n+1}\widehat{\mathcal R}^{R}_{\text{cal}}(a) + \tfrac{B}{n+1} \le \alpha \right\}$
};
\node[output, below=0.9cm of crc] (predict) {
    $\widehat{f}^{R}_{\mathcal{D},\widehat{a}}(x_{n+1})$
};
\node[data, right=1.4cm of predict] (test) {Test $x_{n+1}$};
\draw[arr] (D.south) -- ++(0,-0.35) -| (estimate.north);
\draw[arr] (estimate) -- (invert);
\draw[arr] (invert)   -- (family);
\draw[arr] (family)   -- (crc);
\draw[arr] (crc)      -- (predict);
\draw[arr] (C.east) -- ++(2.4,0) |- (crc.east);
\draw[arr] (test.west) -- (predict.east);
\end{tikzpicture}
\caption{\textbf{The \recirc{} pipeline.} The risk-training set $\cD$ is used
only to train and tune the conditional-risk estimator; its estimated local
risk curves are inverted to define the rectified family
indexed by the risk budget $a$; the calibration set $\cC$ is used only by the
final CRC step, which selects the budget $\widehat a$. Because calibration sees
only observed losses of a monotone one-parameter family, marginal validity
holds regardless of the quality of the estimated curves.}
\label{fig:pipeline}
\end{figure}

The oracle construction uses the unknown curves $R(\lambda\mid x)$. In
practice, we replace them by estimated curves and let CRC calibrate the
remaining scalar risk budget, as summarized in
 the pipeline of
Figure~\ref{fig:pipeline}. The procedure is the following:

\begin{enumerate}
\item Choose an ordered CRC family $\{f_\lambda:\lambda\in\Lam\}$---typically
built from a base predictive model fitted on separate training data---a
bounded loss $\loss\in[0,B]$, and a target marginal risk level $\alpha$.

\item Split the labeled data into a \emph{risk-training} set $\cD$ and a
\emph{calibration} set $\cC=\{(X_j,Y_j)\}_{j=1}^n$.

\item Use $\cD$ exclusively to train and tune the conditional-risk estimator
$\Rhat_{\cD}(\lambda\mid x)$ by directly regressing the loss on
$(x,\lambda)$ (see Section~\ref{sec:estimation} for details). An
alternative based on an estimated law of $Y\mid x$ is described in
Appendix~\ref{app:response-model}.

\item Fix a finite threshold grid
$\Lam_M=\{\lambda_1=\lambda_{\min}<\cdots<\lambda_M=\lambda_{\max}\}$.
For each candidate risk budget $a$, compute
\begin{equation}
  \Rhat_{\cD}^{-1}(a\mid x)
  =
  \min\{\lambda\in\Lam_M:\Rhat_{\cD}(\lambda\mid x)\le a\},
  \label{eq:est-inverse}
\end{equation}
with the conventions $\min\emptyset=\lambda_{\max}$ and
$\Rhat_{\cD}^{-1}(0\mid x)=\lambda_{\max}$. Then define the rectified
decision rules as
\begin{equation*}
  \widehat f^R_{\cD,a}(x)
  =
  f_{\Rhat_{\cD}^{-1}(a\mid x)}(x).
\end{equation*}

\item Fix a finite budget grid
$\mathcal A=\{0=a_0<a_1<\cdots<a_L\}\subset[0,B]$ containing the fully
protective budget $a=0$. On the calibration set, compute
$\widehat{\mathcal R}_{\rm cal}^R(a)
=n^{-1}\sum_{j=1}^n \loss\{\widehat f^R_{\cD,a}(X_j),Y_j\}$ for each
$a\in\mathcal A$. Because $a$ is a risk budget, larger values of $a$ are
\emph{less} protective, so we choose the largest grid budget that still passes
the CRC correction:
\begin{equation}
  \widehat a
  =
  \max\left\{a\in\mathcal A:
  \frac{n}{n+1}\,\widehat{\mathcal R}_{\rm cal}^R(a)
  +
  \frac{B}{n+1}
  \le \alpha
  \right\}.
  \label{eq:ahat}
\end{equation}

\item For a new input $x$, output
$\widehat f^R_{\cD,\widehat a}(x)=f_{\Rhat_{\cD}^{-1}(\widehat a\mid x)}(x)$.
\end{enumerate}

Algorithm~\ref{alg:recirc} in Appendix~\ref{app:algorithms} summarizes the
procedure. The reason the method is valid is simple: after $\cD$ is fixed, the
rules $\{\widehat f^R_{\cD,a}:a\in[0,B]\}$ form a one-parameter family whose
loss is nondecreasing in $a$,  and CRC is applied to this family using only observed calibration losses.
Therefore the standard CRC argument gives
$\E\{\loss(\widehat f^R_{\cD,\widehat a}(X_{n+1}),Y_{n+1})\}\le \alpha$
(Theorem~\ref{thm:marginal-validity}). The guarantee does not require
$\Rhat_{\cD}$ to be correct: poor estimated risk curves can make the
predictions less efficient or less locally adaptive, but they do not invalidate
finite-sample marginal risk control. 

\begin{remark}[Other calibration wrappers]
Rectification is a reparameterization of the family, not of the calibration
scheme. The rectified family $\{\widehat f^R_{\cD,a}\}$ can therefore also be
calibrated with other wrappers, such as risk-controlling prediction sets for
high-probability guarantees \citep{bates2021distribution} or Learn-then-Test
for non-monotone losses \citep{angelopoulos2021learn}. We focus on CRC for
concreteness.
\end{remark}

\subsection{Estimating the local risk curve}
\label{sec:estimation}

The remaining ingredient is the estimator $\Rhat_{\cD}(\lambda\mid x)$. 
Using the risk-training set
$\cD=\{(X_i,Y_i)\}_{i=1}^{m}$, create an augmented loss-training set: for each
observation $i$, sample thresholds $\lambda_{ik}\sim q$, $k=1,\ldots,K$, for a
design distribution $q$ supported on $\Lam$ (e.g., uniform), and compute the
observed losses $Z_{ik}=\loss(f_{\lambda_{ik}}(X_i),Y_i)$. Then train the
conditional-risk regression model
\begin{equation}
  g_\theta(x,\lambda)
  \approx
  \E\{\loss(f_\lambda(X),Y)\mid X=x\}
  =
  R(\lambda\mid x)
  \label{eq:route-r}
\end{equation}
on the augmented data $\{(X_i,\lambda_{ik},Z_{ik})\}$, and set
$\Rhat_{\cD}(\lambda\mid x)=g_\theta(x,\lambda)$. Notice that only a single regression curve is fitted: $x$ and $\lambda$ are the covariates used. The true curve is
non-increasing in $\lambda$, so the fitted surface should respect or
approximate this monotonicity; a simple correction applies isotonic regression
in $\lambda$ for each queried $x$. If the same labeled data are used both to
construct $f_\lambda$ and to fit $g_\theta$, the fitted risk surface may be
optimistically biased; this effect can be reduced by separating the base-model
and risk-training samples, cross-fitting, or using out-of-sample losses.

For computationally heavier settings, we use an alternative implementation in which the risk curve is estimated at a fixed collection of threshold anchors and then interpolated over the full threshold grid, with monotonicity enforced before inversion. The specific experiments and implementation details are reported in Appendix~\ref{app:recirc-implementation}.

An alternative estimator first models $Y\mid x$ and then averages the loss
under that model. Appendix~\ref{app:response-model} gives the construction and
discusses when this generative route may be useful.

\subsection{Special case: rectifying conformal prediction scores}
\label{sec:score-rectification}

Standard conformal prediction with a scalar nonconformity score is a special
case of the framework. Let $S(x,y)$ be a score for which smaller values are
better, and define $f_\lambda(x)=\{y:S(x,y)\le \lambda\}$ with the miscoverage
loss $\loss(f_\lambda(x),y)=\ind\{S(x,y)>\lambda\}$, so that $B=1$. The local
risk curve is the conditional upper tail of the score,
\[
  R(\lambda\mid x)
  =
  \Prob\{S(x,Y)>\lambda\mid X=x\}
  =
  1-F_{S\mid x}(\lambda),
\]
where $F_{S\mid x}$ is the conditional distribution function of $S(x,Y)$ given
$X=x$. The risk budget $a\in[0,1]$ is now a local miscoverage budget, and the
oracle threshold is the conditional quantile
$R^{-1}(a\mid x)=F_{S\mid x}^{-1}(1-a)$, up to the usual generalized-quantile
convention, so the rectified set is
$f_a^R(x)=\{y:S(x,y)\le F_{S\mid x}^{-1}(1-a)\}$. When $F_{S\mid x}$ is
continuous and strictly increasing on the range of the scores, this set can be
rewritten as
\[
  f_a^R(x)
  =
  \{y:\widetilde S(x,y)\le 1-a\},
  \qquad
  \widetilde S(x,y):=F_{S\mid x}(S(x,y)),
\]
which is exactly local score recalibration through the probability integral
transform, as used to obtain approximate conditional coverage in conformal
prediction
(\citealp{chernozhukov2021distributional,izbicki2020flexible,izbicki2022cd}; \citealp{dheur2025unified,cabezas2025epistemic}).

 Appendix~\ref{sec:canonical} extends this quantile interpretation to general
bounded monotone losses: Proposition~\ref{prop:canonical} represents the local
risk curve as the upper tail of a latent variable, so that risk rectification
is a local quantile transformation. This latent variable need not be an
observed conformity score. 

\subsection{A risk-calibration diagnostic for the risk scale}
\label{sec:gof}

 Rectification treats $a$ as a conditional-risk scale. This interpretation can
be assessed on any sample independent of $\cD$, including the calibration set
$\cC$. For a group $G$ of inputs, fixed without using the responses of that
sample, let $I_G=\{i:X_i\in G\}$ and compute, for every budget on the grid,
\begin{equation}
  \widehat r_G(a)
  :=
  \frac{1}{|I_G|}
  \sum_{i\in I_G}
  \loss\left(\widehat f^R_{\cD,a}(X_i),Y_i\right).
  \label{eq:risk-calibration-curve}
\end{equation}
Plotting $\widehat r_G(a)$ against $a$ gives a risk-calibration curve. An exact
risk scale follows the diagonal; a curve below it is conservative; and a
curve above it reveals budgets at which the fitted scale understates risk.
Drawing the curve overall and within prespecified difficulty bins shows both
global mis-scaling and where it occurs. For fixed $a$ and $G$,
$\widehat r_G(a)$ is an unbiased estimate of the group risk bounded in
\eqref{eq:group-conditional-bound}. On $\cC$, the overall curve is the
calibration table already computed by Algorithm~\ref{alg:recirc}, so the
diagnostic requires no extra data.  
Bootstrap or concentration-based bands can be added to the curves. 

Like any grouped diagnostic, \eqref{eq:risk-calibration-curve} assesses risk
on average within the chosen bins rather than at each individual $x$:
deviations of opposite sign can cancel in the overall curve, and finer or
better-chosen bins sharpen the assessment. 

Curves below the diagonal do not necessarily indicate inaccurate risk estimates: even with exact risk curves, discrete prediction sets or limits on how small predictions can become may prevent the risk from reaching the budget.

\section{Theoretical Guarantees}
\label{sec:theory}

This section gives two main guarantees. First, finite-sample marginal risk
control holds regardless of the quality of the estimated risk curves. Second,
accurate curves turn the budget into an approximate conditional-risk scale;
when the curves cross the target, the deployed conditional risk approaches
$\alpha$. Most-input and group versions of the second guarantee, and a more
general form of its asymptotic version, are shown in
Appendix~\ref{app:representations}; all proofs are in
Appendix~\ref{app:proofs}.

\subsection{Finite-sample marginal risk control}
\label{sec:marginal-validity}

Recall that $\cD$ is the risk-training set, 
$\cC=\{(X_j,Y_j)\}_{j=1}^n$ the calibration set, and $(X_{n+1},Y_{n+1})$ the
test point. To avoid clutter, conditioning on $\cD$ below also conditions on
any independent data and randomness used to fit the base family or the risk
estimator before calibration.  We will, when necessary, assume the following:

\begin{assumption}[Exchangeability]
\label{ass:exchange}
$(X_1,Y_1),\ldots,(X_{n+1},Y_{n+1})$ are exchangeable and independent of
$\cD$.
\end{assumption}

\begin{assumption}[Monotone bounded family]
\label{ass:monotone}
The loss satisfies $\loss(f_\lambda(x),y)\in[0,B]$ and
\eqref{eq:monotone-family}, and the family contains a fully protective rule:
$\loss(f_{\lambda_{\max}}(x),y)=0$ for every $(x,y)$.
\end{assumption}

\begin{assumption}[Pre-calibration finite grid]
\label{ass:regularity}
The budget in \eqref{eq:ahat} is selected from a finite grid
$\mathcal A=\{0=a_0<a_1<\cdots<a_L\}\subset[0,B]$ containing $a_0=0$.
\end{assumption}

Assumptions~\ref{ass:exchange} and~\ref{ass:monotone} are standard in CRC.
The former is its distributional requirement; for the latter, the full label
set, the full mask, or the whole response space incurs zero false-negative
loss. Assumption~\ref{ass:regularity} simply formalizes how the
procedure is implemented (Algorithm~\ref{alg:recirc}).

\begin{theorem}[Marginal validity of \recirc{}]
\label{thm:marginal-validity}
Fix $0<\alpha\le B$. Suppose
Assumptions~\ref{ass:exchange}--\ref{ass:regularity} hold and
$n\ge B/\alpha-1$. Then the budget $\widehat a$ in \eqref{eq:ahat} is well
defined and
\[
  \E\left\{\loss\left(\widehat f^R_{\cD,\widehat a}(X_{n+1}),Y_{n+1}\right)
  \,\middle|\, \cD\right\}
  \le \alpha .
\]
In particular, the unconditional risk is also at most $\alpha$.
\end{theorem}

Theorem~\ref{thm:marginal-validity} states that rectification preserves the
finite-sample marginal guarantee of CRC, no matter how poor the estimated risk
curves are.

\subsection{Conditional risk control}
\label{sec:conditional}

We now quantify in what sense the budget $a$ controls conditional risk.
First, we show that the oracle budget is a conditional-risk upper bound.

\begin{proposition}[Oracle conditional validity]
\label{prop:oracle-conditional}
Fix $x$. Suppose Assumption~\ref{ass:monotone} holds and
$R(\cdot\mid x)$ is right-continuous. Then, for every $a\in[0,B]$,
$
  \E\{\loss(f_a^R(X),Y)\mid X=x\}
   \le a.
$
If in addition $R(\cdot\mid x)$ is continuous and $a\le R(\lambda_{\min}\mid
x)$, the inequality holds with equality.
\end{proposition}

The equality condition says that the least protective rule must itself have
risk at least as large as the requested budget. If it does not, the local
scale stops at $R(\lambda_{\min}\mid x)$; the upper bound still holds, but
equality cannot.

In practice the curves are estimated, and the oracle bound degrades by exactly
the amount by which the fitted curve understates the true risk:

\begin{proposition}[Conditional validity from risk-curve accuracy]
\label{prop:approx-conditional}
Fix $x$ and suppose Assumptions~\ref{ass:exchange} and~\ref{ass:monotone}
hold. Define
\[
  \varepsilon_M(x)
  :=
  \max_{\lambda\in\Lam_M}
  [R(\lambda\mid x)-\Rhat_\cD(\lambda\mid x)]_+.
\]
Then, simultaneously for every $a\in[0,B]$,
\[
  \E\left\{\loss\left(\widehat f^R_{\cD,a}(X),Y\right)
  \,\middle|\,X=x,\cD\right\}
  \le a+\varepsilon_M(x).
\]
If, in addition, the calibration and test pairs are i.i.d.,
Assumption~\ref{ass:regularity} holds, $0<\alpha\le B$, and
$n\ge B/\alpha-1$, the bound remains valid after conditioning on the
calibration set and replacing $a$ by $\widehat a$.
\end{proposition}

At the calibrated budget, the proposition controls risk relative to
$\widehat a$ rather than relative to $\alpha$ itself, and the two can differ
when the rectified family is conservative or saturates at some inputs.
Proposition~\ref{prop:aggregate-conditional} in
Appendix~\ref{app:aggregate-conditional} converts the pointwise bound into
``most-input'' and arbitrary-group statements. 
Proposition~\ref{prop:risk-rate} in Appendix~\ref{app:representations}
quantifies how risk-curve estimation error and threshold-grid spacing
translate into error in the conditional-risk scale over attainable budgets. 

The next result closes the
remaining gap, giving conditions under which $\widehat a$, and with it the
deployed conditional risk, converges to the target level.

\begin{corollary}[Rate for the deployed conditional risk; corollary of
Theorem~\ref{thm:deployed-conditional}]
\label{cor:deployed-conditional}
Suppose Assumptions~\ref{ass:exchange}--\ref{ass:regularity} hold, with the
calibration and test pairs i.i.d. Along a sequence $n\to\infty$, let the
risk-training size satisfy $m=m(n)\to\infty$, let $M=M(n)$, and
 let the threshold grid have maximum
spacing $h_M\to0$, and let the budget grid
$\mathcal A_n=\{0=a_{0,n}<\cdots<a_{L_n,n}=B\}$  with maximum
spacing $h_n\to0$. 
All quantities below are indexed along this sequence: $\eta_m$ and $h_M$
abbreviate $\eta_{m(n)}$ and the spacing of $\Lam_{M(n)}$, and every
$O_{\Prob}$ statement is taken as $n\to\infty$.
Fix $\alpha\in(0,B)$. Assume there is a constant $c>0$ such that
$R(\lambda_{\min}\mid x)\ge\alpha+c
$ 
for almost every $x$. Suppose also that the curves $R(\cdot\mid x)$ share a
finite Lipschitz constant and that, uniformly over inputs and grid thresholds,
their estimation error is $O_{\Prob}(\eta_m)$ for some $\eta_m\to0$; that is,
\[
  \operatorname*{ess\,sup}_{x\sim P_X}
  \max_{\lambda\in\Lam_M}
  |R(\lambda\mid x)-\Rhat_{\cD_m}(\lambda\mid x)|
  =O_{\Prob}(\eta_m).
\]
Then, provided $n\ge B/\alpha-1$, for almost every fixed $x$,
\[
  \left|
  \E\left\{\loss\left(\widehat f^R_{\cD_m,\widehat a}(X),Y\right)
  \,\middle|\,X=x,\cD_m,\cC\right\}-\alpha
  \right|
  =O_{\Prob}(\eta_m+h_M+n^{-1/2}+h_n).
\]
\end{corollary}

The endpoint condition in the corollary ensures that risk $\alpha$ is attainable for almost
every input. The rate combines errors from learning the risk curves,
discretizing the threshold and budget, and using a finite calibration sample.
Here $\eta_m$ is a standard uniform regression rate for the risk surface
$(x,\lambda)\mapsto R(\lambda\mid x)$. For example, if $x$ is $d$-dimensional
and this surface is $s$-H\"older smooth,
local-polynomial or spline/wavelet regression can attain
\[
  \eta_m=(\log m/m)^{s/(2s+d+1)}
\]
under the usual design, moment, and tuning conditions
\citep{stone1982optimal,masry1996multivariate,chen2015optimal}; the extra one
in the denominator comes from using $\lambda$ as an additional predictor.
Theorem~\ref{thm:deployed-conditional} in
Appendix~\ref{app:representations} states the general version, in which the
marginal and local risk-scale errors are allowed to have different rates and
no uniformity over $x$ is required.

The pointwise guarantee in Proposition~\ref{prop:approx-conditional} also has
a deployment implication: it remains valid if deployment only reweights the
inputs. Formally,  covariate shift means that the input distribution
changes from $P_X$ during calibration to $Q_X$ at deployment, while the
response mechanism $Y\mid X$ remains unchanged \citep{shimodaira2000improving,izbicki2017photoz}. 

\begin{corollary}[Estimated \recirc{} under covariate shift]
\label{cor:shift}
Fix $\cD$ and $\cC$, and let
$\widehat f=\widehat f^R_{\cD,\widehat a}$ be the deployed rule. If
$Q_{Y\mid X}=P_{Y\mid X}$ for $P_X$-almost every $x$ and $Q_X\ll P_X$, then,
under the conditions of Proposition~\ref{prop:approx-conditional},
\begin{equation}
  \E_Q\{\loss(\widehat f(X),Y)\}-\widehat a
  \le \|\varepsilon_M\|_{L^\infty(P_X)}.
  \label{eq:shift-estimated}
\end{equation}
\end{corollary}

Thus the worst local underestimation of the fitted risk curves controls
deployed risk under any such shift.

\section{Experiments}
\label{sec:experiments}

We evaluate \recirc{} across eight synthetic and real-data settings designed to capture different forms of risk heterogeneity. We examine whether rectification reduces groupwise risk heterogeneity while maintaining marginal risk close to the target level, and characterize the associated changes in prediction size. The synthetic experiments provide controlled mechanisms of heterogeneity, whereas the real-data applications span image segmentation, multilabel text classification, multiclass prediction, and tabular regression. Within each setting, we compare global CRC, AA-CRC, and \recirc{}, with the latter using TabICLv2 as a regression model to estimate the conditional-risk surface \citep{qu2026tabiclv2}.

\subsection{Experimental setup}
\label{sec:exp-protocol}

Detailed experimental settings, including base predictors, adaptive representations, sample sizes, and data splits, are reported in Appendix~\ref{app:experimental-settings}. The three synthetic designs introduce distinct sources of heterogeneity. The first is a heteroscedastic regression problem with $X\sim\mathrm{Unif}(-2,2)$ and $Y=\sin(\pi X/2)+\sigma(X)\varepsilon$, where $\varepsilon\sim N(0,1)$ and $\sigma(x)=0.2+0.6|x|$. The second is a multilabel problem in which a one-dimensional difficulty variable controls both label cardinality and the separation between positive and negative scores. The third introduces interaction effects through an XOR structure, with conditional scale $\sigma(x)=0.4+2.2\,\ind\{\operatorname{sign}(x_1)\neq\operatorname{sign}(x_2)\}$. These controlled settings are complemented by five real-data benchmarks spanning image segmentation, text classification, multiclass prediction, and tabular regression: polyp segmentation uses PraNet pixelwise scores; RCV1 and Letter Recognition construct prediction sets from probabilistic classifiers; and Medical Insurance and Superconductor use quantile random forests  \citep{meinshausen2006quantile} to construct regression intervals.

\paragraph{Data allocation and repetitions.}

As in Section~\ref{sec:procedure}, \recirc{} uses a risk-training set $\cD$ to fit the conditional-risk estimator and an independent calibration set $\cC$ for the final CRC step. Within each repetition, all methods share the same base predictor, nested decision family, and test observations, so their comparisons are paired. Because global CRC does not require risk training, it is calibrated on $\cD\cup\cC$ in the heteroscedastic, latent-difficulty, and RCV1 settings. In the QRF-based settings, $\cD$ is used to fit the base model, so global CRC is calibrated on $\cC$ alone; in polyp segmentation and Letter Recognition, it uses the same calibration set $\cC$ as \recirc{}. The size of the calibration sample affects only the variability of the global threshold, not its inability to adapt to input difficulty, which drives the groupwise comparisons below. AA-CRC follows its original data allocation except in polyp segmentation and Letter Recognition (Appendix~\ref{app:aacrc-implementation}).

The QRF-based settings require additional care because the base quantile random forest is fitted on $\cD$. If ordinary fitted predictions were used on the same observations to construct the \recirc{} risk-training data, the resulting losses could be optimistically biased. We therefore compute the QRF-derived features and losses for observations in $\cD$ from out-of-bag quantile predictions, so that each observation is evaluated using only trees for which it was out of bag. Calibration and test observations use ordinary predictions from the fitted QRF. The target risk level is $\alpha=0.10$ throughout, and all results are averaged over $20$ repetitions.

\paragraph{Methods.}

Global CRC calibrates a single threshold shared by all inputs. AA-CRC \citep{blot2024automatically} instead learns an input-dependent threshold within a chosen adaptive function class, using the objective and gradient of the authors' implementation. In the RF-based settings, we fit a random forest on $\cD$ to a task-specific measure of prediction difficulty and use its leaf-membership indicators as the feature map $\Phi(x)$. The resulting threshold has the form
\[
u_\theta(x)=\Phi(x)^\top\theta,
\]
with the coefficients fitted on the independent calibration set $\cC$. Because each tree contributes exactly one active leaf indicator, constant functions already belong to the span of $\Phi(x)$, so no additional intercept is needed.

AA-CRC and \recirc{} use covariate information differently. AA-CRC adapts the decision parameter directly through the function $u_\theta(x)$. \recirc{} estimates the conditional-risk surface $R(\lambda\mid x)$ as a function of an input representation and $\lambda$, and then inverts the fitted surface to obtain an input-specific threshold for each risk budget. The final scalar budget is calibrated by CRC.

The representations used by the two methods are therefore task-specific and need not coincide. In the heteroscedastic experiment, AA-CRC learns its representation from the covariate and absolute residuals, whereas \recirc{} estimates risk jointly from the covariate and $\lambda$. In the synthetic multilabel experiment, both methods use the difficulty covariate and score summaries, with \recirc{} additionally conditioning on $\lambda$. In the QRF-based settings, we follow the random-forest representation proposed for tabular data in AA-CRC \citep{blot2024automatically}: AA-CRC learns an RF representation from the tabular covariates and a minimum-covering-multiplier target, whereas \recirc{} also uses QRF-derived location and scale summaries to estimate the risk surface.

\recirc{} uses TabICLv2 to estimate the local risk curves. Depending on the setting, we either fit a single regression model jointly on the input representation and $\lambda$, or fit separate regressors at a collection of threshold anchors and interpolate between the estimated risks, enforcing monotonicity before inversion. For complex inputs such as images, the risk estimator receives compact summaries from the base model rather than the raw covariates. These representations need not recover all information relevant to the conditional loss: as discussed in Section~\ref{sec:procedure}, an imperfect risk model may reduce local adaptivity but does not affect the finite-sample marginal guarantee.

Appendix~\ref{app:experimental-settings} gives the complete specification for each experiment, including the base-model fitting sample, nested decision family, threshold and budget grids, risk-regression representation, threshold-sampling or anchor construction, AA-CRC regularization, and evaluation groups.

\subsection{Losses and evaluation metrics}
\label{sec:exp-metrics}

Each experiment uses a bounded loss appropriate to its task. For multilabel prediction and segmentation, we use the missed-positive loss  $\loss_{\rm FN}$ of Section~\ref{sec:crc}, that is, the fraction of positive labels or pixels omitted from the prediction. The corresponding risk is the per-observation average of this loss; in segmentation, this gives equal weight to images rather than pooling positive pixels across the dataset. Letter Recognition uses the miscoverage loss
$$
\loss_{\mathrm{mc}}(f_\lambda(x),y)
=
\ind\{y\notin f_\lambda(x)\}.
$$
The heteroscedastic regression experiment uses symmetric intervals
$[\widehat\mu(x)-\lambda,\widehat\mu(x)+\lambda]$, where $\widehat\mu(x)$ is the fitted GAM mean, together with the bounded excess loss
$$
\loss_{\mathrm{exc}}(f_\lambda(x),y)
=
\min\left\{
1,\frac{(|y-\widehat\mu(x)|-\lambda)_+}{1.5}
\right\}.
$$
The QRF-based regression experiments use intervals
$[l_\lambda(x),u_\lambda(x)]$ centered at the estimated conditional median, together with the asymmetric miscoverage loss
$$
\loss_{\mathrm{asym}}(f_\lambda(x),y)
=
0.2\,\ind\{y<l_\lambda(x)\}
+
0.8\,\ind\{y>u_\lambda(x)\}.
$$
Since a response cannot fall below and above the interval simultaneously, this loss is bounded by $0.8$. The CRC correction uses this bound in Superconductor and the conservative bound $B=1$ in Medical Insurance and XOR; both are valid, since $B$ only needs to upper-bound the loss.

For repetition $r$ and method $j$, let $T_r$ denote the test set and
$L_{rji}$ the observed loss on test observation $i\in T_r$. The empirical
marginal risk is
\[
  \bar L_{rj}
  =
  \frac{1}{|T_r|}
  \sum_{i\in T_r}L_{rji}.
\]
To assess heterogeneity, let
$\mathcal G_r=\{G_{r1},\ldots,G_{rJ_r}\}$ denote the collection of evaluation
groups, common to all methods within a repetition. Groups with too few test
observations are omitted (Appendix~\ref{app:evaluation-groups}), so $J_r$ may
vary across repetitions. The empirical risk in group $k$ is
\[
  \bar L_{rjk}
  =
  \frac{1}{|T_r\cap G_{rk}|}
  \sum_{i\in T_r\cap G_{rk}}L_{rji},
\]
and we summarize groupwise performance by
\[
  W_{rj}
  =
  \max_{k\le J_r}\bar L_{rjk},
  \qquad
  E_{rj}
  =
  \frac{1}{J_r}
  \sum_{k=1}^{J_r}
  (\bar L_{rjk}-\alpha)_+.
\]

Here, $W_{rj}$ is the worst-group risk and $E_{rj}$ is the mean positive group excess above the target. The latter gives equal weight to groups, even when their sizes differ or the groups overlap.

Groups are defined using setting-specific measures of prediction difficulty and are held fixed across methods within each repetition. The synthetic experiments use the known heterogeneity structure, while the real-data benchmarks use predictive uncertainty or task-relevant covariates to construct the groups. Detailed group definitions are given in Appendix~\ref{app:evaluation-groups}. The Superconductor groups were selected using the responses, so this grouped analysis is exploratory. 

Prediction size is measured as the number of selected labels, classes, or pixels, or as interval width in regression. Tables~\ref{tab:worst-risk}--\ref{tab:marginal-risk} report the averages across the 20 repetitions of worst-group risk, mean positive group excess, prediction size, and marginal risk. Worst-group risk is computed separately within each repetition before averaging. These summaries measure heterogeneity across the chosen groups rather than pointwise conditional risk.

\subsection{Results}
\label{sec:exp-results}

Across all eight settings, \recirc{} attains the lowest average worst-group risk and mean positive group excess, although in some settings its advantage over AA-CRC is within sampling variability, while its effect on prediction size varies across applications (Tables~\ref{tab:worst-risk}--\ref{tab:marginal-risk}).

\paragraph{Groupwise risk.}
The synthetic experiments show consistent reductions in risk heterogeneity. In the heteroscedastic setting, \recirc{} attains an average worst-group risk of $0.1130$, compared with $0.1312$ for AA-CRC and $0.2340$ for global CRC. In the latent-difficulty setting, the corresponding values are $0.1179$, $0.1303$, and $0.2329$. The mean positive group excess is also smallest for \recirc{} in both settings. This ordering persists across distinct forms of heterogeneity: the first two designs involve one-dimensional difficulty mechanisms, whereas the XOR experiment introduces interaction-driven heterogeneity. In the latter, \recirc{} attains an average worst-group risk of $0.1130$, compared with $0.1166$ for AA-CRC and $0.1338$ for global CRC, with corresponding mean positive group excesses of $0.0041$, $0.0066$, and $0.0145$.

The same pattern extends to the five real-data benchmarks, where \recirc{} attains the smallest average worst-group risk and mean positive group excess in every setting. The differences are particularly pronounced in RCV1, where worst-group risk decreases from $0.1722$ under global CRC and $0.1410$ under AA-CRC to $0.1144$ under \recirc{}, and in Medical Insurance, where the corresponding values are $0.2021$, $0.2061$, and $0.1283$; in the latter, AA-CRC does not improve on global CRC. In polyp segmentation and Letter Recognition, the reductions relative to global CRC are also large (from $0.3941$ to $0.1092$ and from $0.2861$ to $0.1848$), whereas the improvements over AA-CRC are smaller; in Superconductor, differences among the three methods are modest. In Letter Recognition, the worst-group risk remains well above $\alpha$ for all methods: because every top-$m$ set contains at least one class, easy inputs cannot reach the target risk, which pushes $\widehat a$ above $\alpha$ (Proposition~\ref{prop:approx-conditional}).

\paragraph{Marginal risk.}
 The groupwise comparisons should be interpreted together with the marginal operating level of each method (Table~\ref{tab:marginal-risk}). The CRC guarantee concerns expected loss over calibration and a new observation and therefore does not require empirical test risk to remain below $\alpha$ in every repetition. Across the eight experiments, the mean marginal risk of \recirc{} ranges from $0.0908$ to $0.0985$, remaining close to the target $\alpha=0.10$. Global CRC also operates slightly below the target on average, while AA-CRC remains close to the target but slightly exceeds $0.10$ in some settings. Relative to global CRC, which operates at a similar marginal level, the reductions in groupwise risk achieved by \recirc{} are therefore not explained by a lower marginal risk. Relative to AA-CRC, part of the advantage in the XOR and Superconductor settings is of the same order as the difference in marginal risk.

\begin{figure}[t!]
    \centering
    \includegraphics[width=\linewidth]{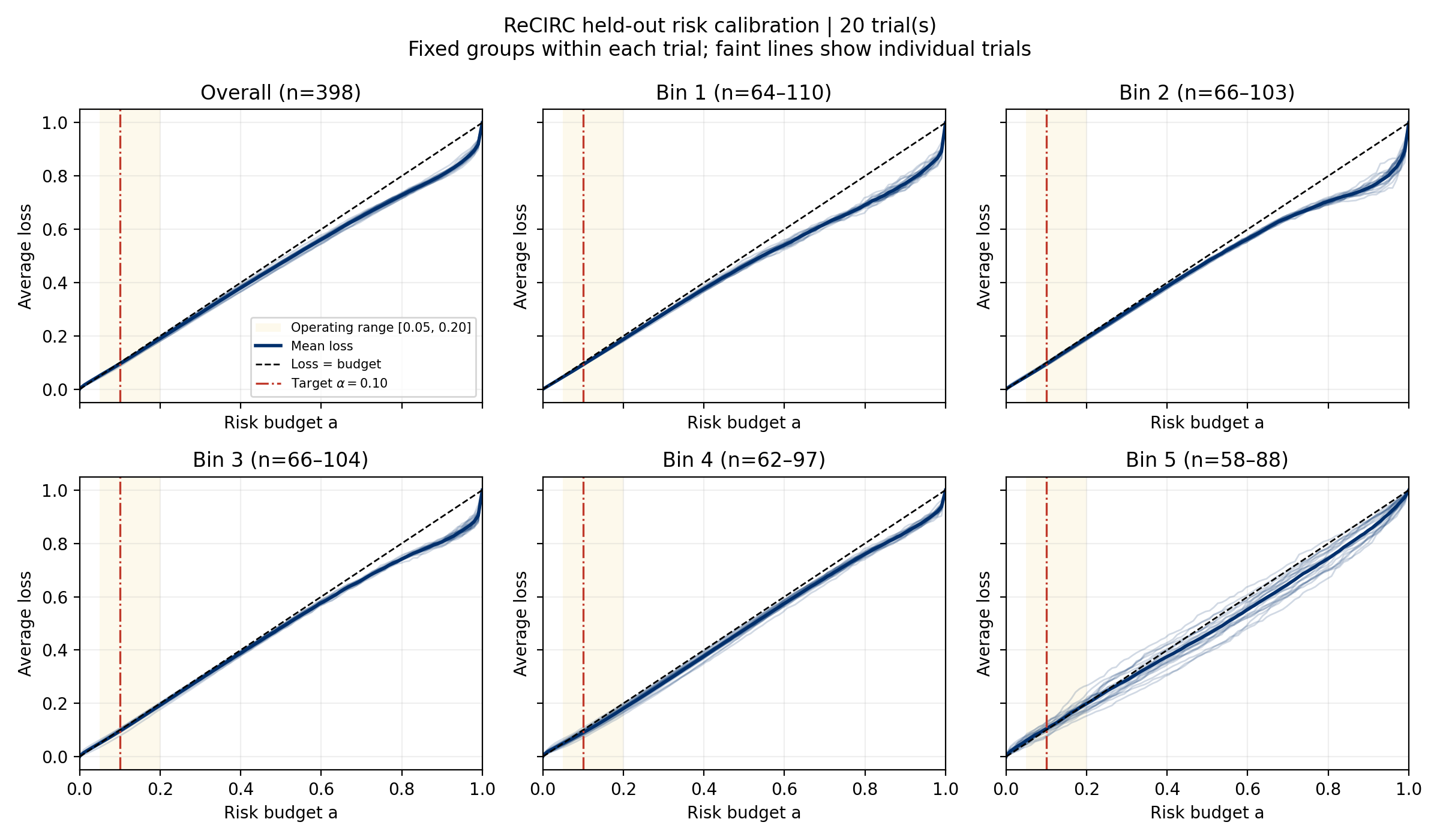}
    \caption{Held-out risk-calibration diagnostics for \recirc{} in the polyp-segmentation experiment. Each panel plots empirical average loss $\widehat r_G(a)$ against the risk budget $a$, overall or within one of five groups defined by a label-free uncertainty score. Loss is the fraction of positive pixels omitted, averaged across images. Thick blue curves show means
    over 20 repetitions; faint curves show individual repetitions, with groups fixed within each repetition. The dashed black diagonal marks equality between risk and budget, the red dash-dotted vertical line marks $\alpha=0.10$, and the shaded
    band highlights $a\in[0.05,0.20]$.}
    \label{fig:risk-calibration-3}
\end{figure}

\paragraph{Risk-scale calibration.}
The risk-calibration diagnostics of Section~\ref{sec:gof} are computed on the test set of each repetition, which is used neither to fit the risk curves nor to calibrate the budget. Figure~\ref{fig:risk-calibration-3} shows the polyp-segmentation experiment. The empirical risk curves approximately follow the
diagonal over $a\in[0.05,0.20]$, both overall and within the five uncertainty groups, supporting the interpretation of $a$ as a common group-level risk scale near the target $\alpha=0.10$. At larger budgets, the mean curves generally lie below the
diagonal, indicating conservative risk relative to the budget. 
The remaining diagnostics, reported in Appendix~\ref{app:calibration-diagnostics}, show similar behavior near $\alpha$, except in the low-entropy Letter Recognition groups; at larger budgets, their departures from the diagonal are mostly conservative and reflect endpoint saturation, with the regression intervals collapsing to a point and the Letter Recognition sets reducing to a single class.

\paragraph{Prediction size.} The reduction in risk heterogeneity can have different consequences for prediction size (Table~\ref{tab:set-size}).  In polyp segmentation, the decrease in worst-group risk relative to global CRC is accompanied by an increase in average mask area from $15{,}538$ to $22{,}455$ pixels, approximately $45\%$. In RCV1, average prediction-set size increases from $5.016$ to $15.392$ labels, and smaller increases occur in the latent-difficulty and XOR experiments. AA-CRC shows increases of similar magnitude in polyp segmentation and RCV1 ($20{,}258$ pixels and $13.860$ labels), which suggests that this cost reflects the additional protection given to difficult inputs rather than rectification specifically. 

In other settings, however, improved groupwise risk is obtained together with smaller predictions. Relative to global CRC, \recirc{} reduces average set size from $3.100$ to $2.210$ in Letter Recognition, and average interval width by approximately $29\%$ in Medical Insurance and $12\%$ in Superconductor. In the heteroscedastic experiment, \recirc{} also produces slightly shorter intervals than global CRC, although AA-CRC attains the smallest average width. These results show that the size cost of redistributing risk across inputs is strongly application-dependent: in some settings more uniform risk requires substantially larger predictions, whereas in others adaptation allows both lower groupwise risk and smaller average predictions.

\providecommand{\singlespacing}{\renewcommand{\baselinestretch}{1}\selectfont}
\begin{table}[H]
\centering
\singlespacing
\caption{Worst-group risk at $\alpha=0.10$. Entries report mean values across repetitions, with standard errors in parentheses. Bold indicates the lowest mean within each setting.}
\label{tab:worst-risk}
\footnotesize
\setlength{\tabcolsep}{4.5pt}
\renewcommand{\arraystretch}{0.95}

\begin{tabular}{@{}lccc@{}}
\toprule
Setting
&  Global CRC
& AA-CRC
& \recirc{}--TabICLv2 \\
\midrule
\multicolumn{4}{@{}l}{\emph{Synthetic settings}} \\

Heteroscedastic
& 0.2340 (0.0061)
& 0.1312 (0.0047)
& \textbf{0.1130 (0.0039)} \\

Latent difficulty
& 0.2329 (0.0044)
& 0.1303 (0.0038)
& \textbf{0.1179 (0.0022)} \\

XOR interaction
& 0.1338 (0.0026)
& 0.1166 (0.0021)
& \textbf{0.1130 (0.0016)} \\

\addlinespace[2pt]
\multicolumn{4}{@{}l}{\emph{Real benchmarks}} \\

Polyp segmentation
& 0.3941 (0.0135)
& 0.1148 (0.0042)
& \textbf{0.1092 (0.0027)} \\

RCV1
& 0.1722 (0.0018)
& 0.1410 (0.0021)
& \textbf{0.1144 (0.0013)} \\

Letter Recognition
& 0.2861 (0.0054)
& 0.1937 (0.0026)
& \textbf{0.1848 (0.0038)} \\

Medical Insurance
& 0.2021 (0.0168)
& 0.2061 (0.0190)
& \textbf{0.1283 (0.0065)} \\

Superconductor
& 0.1462 (0.0096)
& 0.1371 (0.0068)
& \textbf{0.1325 (0.0099)} \\

\bottomrule
\end{tabular}
\end{table}

\begin{table}[H]
\centering
\singlespacing
\caption{Mean positive group excess  at $\alpha=0.10$.
Entries report mean values across repetitions, with standard errors in parentheses. Bold indicates the lowest mean within each setting.}
\label{tab:excess-risk}
\footnotesize
\setlength{\tabcolsep}{4.5pt}
\renewcommand{\arraystretch}{0.95}

\begin{tabular}{@{}lccc@{}}
\toprule
Setting
&  Global CRC
& AA-CRC
& \recirc{}--TabICLv2 \\
\midrule
\multicolumn{4}{@{}l}{\emph{Synthetic settings}} \\

Heteroscedastic
& 0.0371 (0.0019)
& 0.0083 (0.0010)
& \textbf{0.0043 (0.0010)} \\

Latent difficulty
& 0.0331 (0.0012)
& 0.0093 (0.0013)
& \textbf{0.0058 (0.0011)} \\

XOR interaction
& 0.0145 (0.0012)
& 0.0066 (0.0008)
& \textbf{0.0041 (0.0006)} \\

\addlinespace[2pt]
\multicolumn{4}{@{}l}{\emph{Real benchmarks}} \\

Polyp segmentation
& 0.0588 (0.0027)
& 0.0032 (0.0008)
& \textbf{0.0027 (0.0006)} \\

RCV1
& 0.0311 (0.0009)
& 0.0129 (0.0009)
& \textbf{0.0032 (0.0003)} \\

Letter Recognition
& 0.0413 (0.0013)
& 0.0320 (0.0009)
& \textbf{0.0292 (0.0012)} \\

Medical Insurance
& 0.0243 (0.0049)
& 0.0286 (0.0057)
& \textbf{0.0099 (0.0024)} \\

Superconductor
& 0.0128 (0.0024)
& 0.0108 (0.0018)
& \textbf{0.0093 (0.0027)} \\
\bottomrule
\end{tabular}
\end{table}

\begin{table}[H]
\centering
\singlespacing
\caption{Mean prediction size or interval width at $\alpha=0.10$. Entries report mean values across repetitions, with standard errors in parentheses. Bold indicates the lowest mean within each setting.}
\label{tab:set-size}
\footnotesize
\setlength{\tabcolsep}{4.5pt}
\renewcommand{\arraystretch}{0.95}

\begin{tabular}{@{}lccc@{}}
\toprule
Setting
& Global CRC
& AA-CRC
& \recirc{}--TabICLv2 \\
\midrule
\multicolumn{4}{@{}l}{\emph{Synthetic settings}} \\

Heteroscedastic
& 1.764 (0.018)
& \textbf{1.615 (0.021)}
& 1.660 (0.022) \\

Latent difficulty
& \textbf{5.380 (0.030)}
& 6.528 (0.055)
& 6.813 (0.044) \\

XOR interaction
& \textbf{3.948 (0.027)}
& 4.000 (0.023)
& 4.065 (0.016) \\

\addlinespace[2pt]
\multicolumn{4}{@{}l}{\emph{Real benchmarks}} \\

Polyp segmentation
& \textbf{15{,}538 (187)}
& 20{,}258 (195)
& 22{,}455 (335) \\

RCV1
& \textbf{5.016 (0.027)}
& 13.860 (0.194)
& 15.392 (0.242) \\

Letter Recognition
& 3.100 (0.069)
& 6.060 (0.046)
& \textbf{2.210 (0.022)} \\

Medical Insurance
& 9{,}198.6 (282.2)
& 8{,}459.4 (221.6)
& \textbf{6{,}558.5 (210.3)} \\

Superconductor
& 28.352 (0.214)
& 26.781 (0.102)
& \textbf{24.865 (0.119)} \\

\bottomrule
\end{tabular}
\end{table}

\begin{table}[H]
\centering
\singlespacing
\caption{Marginal risk at $\alpha=0.10$. Entries report mean values across repetitions, with standard errors in parentheses.}
\label{tab:marginal-risk}
\footnotesize
\setlength{\tabcolsep}{4.5pt}
\renewcommand{\arraystretch}{0.95}

\begin{tabular}{@{}lccc@{}}
\toprule
Setting
& Global CRC
& AA-CRC
& \recirc{}--TabICLv2 \\
\midrule
\multicolumn{4}{@{}l}{\emph{Synthetic settings}} \\

Heteroscedastic
& 0.0963 (0.0022)
& 0.0975 (0.0017)
& 0.0908 (0.0021) \\

Latent difficulty
& 0.0924 (0.0017)
& 0.0993 (0.0023)
& 0.0941 (0.0021) \\

XOR interaction
& 0.0937 (0.0014)
& 0.1018 (0.0009)
& 0.0975 (0.0008) \\

\addlinespace[2pt]
\multicolumn{4}{@{}l}{\emph{Real benchmarks}} \\

Polyp segmentation
& 0.0917 (0.0027)
& 0.0848 (0.0010)
& 0.0956 (0.0014) \\

RCV1
& 0.0989 (0.0010)
& 0.1075 (0.0011)
& 0.0952 (0.0010) \\

Letter Recognition
& 0.0952 (0.0017)
& 0.0983 (0.0012)
& 0.0985 (0.0014) \\

Medical Insurance
& 0.0917 (0.0041)
& 0.0990 (0.0044)
& 0.0936 (0.0041) \\

Superconductor
& 0.0903 (0.0016)
& 0.1002 (0.0009)
& 0.0967 (0.0011) \\

\bottomrule
\end{tabular}
\end{table}

\section{Final Remarks}
\label{sec:final-remarks}

This paper discusses the unevenness of conformal risk control across the
covariate space: the parameter calibrated by CRC is a raw threshold, so a
 single global value can overprotect easy inputs and underprotect hard ones.  We
proposed \recirc{}, a framework that reparameterizes the calibrated family so
that its parameter becomes a local risk budget, obtained by inverting
estimated local risk curves, and then applies ordinary CRC unchanged. The
framework directly regresses observed losses on the input and threshold and
 retains the finite-sample marginal guarantee of CRC regardless of the
quality of the fitted curves. Accurate curves yield approximate conditional
risk control, while a direct risk-calibration curve provides a goodness-of-fit
diagnostic for the estimated risk scale. Under  uniformly consistent risk-curve estimates, a target attainable at almost every input, finer threshold and budget grids, and a growing
calibration sample, the deployed conditional risk converges to the target.

Two consequences of learning the family are
worth emphasizing. First, the fitted risk surface does not depend on the
target level: a single fit of $\Rhat_{\cD}(\lambda\mid x)$ serves every
$\alpha$, since changing the target only re-scans the calibration table of
Algorithm~\ref{alg:recirc}, and because the rectified loss is monotone in the
budget, the resulting predictors are automatically nested across targets. Methods whose fitting objective is indexed by
$\alpha$, including AA-CRC as formulated here, generally require a separate
fit at each level and do not automatically guarantee nesting across levels.  Second, the
risk-calibration curve of Section~\ref{sec:gof} is  possible 
because $a$ is an advertised conditional-risk level that can be confronted
 with realized loss; this has no analogue when the calibrated
parameter is a raw threshold as in the standard formulation of CRC. 

Despite offering stronger local risk behavior, \recirc{} has some limitations.
First, it requires an additional risk-training split to estimate the risk
curves, which reduces the data available for the base model and for
calibration. 
Second, the augmented loss-regression data add computation. Third, the finite-sample guarantee remains marginal: the local
improvements depend on the quality of the estimated risk curves, and flat or
discontinuous local risk curves can make the calibrated budget conservative at
some inputs. Fourth, more uniform risk can require larger predictions: in polyp segmentation and RCV1, average prediction size increased substantially relative to global CRC, although in  four of the other six settings it decreased.

This work opens several avenues for future research. A first direction is to
remove the dedicated risk-training split through cross-fitting or
cross-validation-style calibration arguments, which would eliminate
rectification's additional data requirement relative to standard CRC. A second direction is to combine
rectification with partition-based calibration, in the spirit of Mondrian
conformal methods \citep{vovk2012conditional,cabezas2025regression}, so as to
 obtain finite-sample local guarantees on the rectified scale; the price
such methods pay in multiplicity may be lower here precisely because the
rectified risk profile is already close to flat.  

A third direction concerns the calibration wrapper itself. Because
rectification acts on the decision family rather than on the calibration step,
it composes with any procedure that calibrates a scalar along an ordered
family: handing $\{\widehat f^R_{\cD,a}\}$ to risk-controlling prediction sets
\citep{bates2021distribution} or to Learn-then-Test
\citep{angelopoulos2021learn} requires no modification of either, and would
deliver high-probability rather than in-expectation control over a budget that
retains its local interpretation. The existing literature reaches non-marginal
guarantees under these wrappers by a different route, refining \emph{where}
calibration happens: separate thresholds within a prespecified partition or
hierarchy, or a reweighting class fixed in advance. Rectification  is complementary, and in a useful sense orthogonal.

In conclusion, by turning the knob calibrated by CRC into a common conditional-risk scale,
\recirc{} makes distribution-free risk control locally meaningful while
keeping its guarantees intact.
 Code to implement \recirc{} and reproduce the experiments is available at \href{https://github.com/heltongraziadei/ReCIRC}{repository on Github}.

\section{Disclosure statement}
\label{sec:disclosure}

The authors report there are no competing interests to declare.

\section{Data Availability Statement}

Reproduction code for all experiments is provided as Supplementary Material and is also publicly available at \href{https://github.com/heltongraziadei/ReCIRC}{repository on Github}. The three synthetic settings described in Section~\ref{sec:exp-protocol} are generated directly by the corresponding scripts using predefined random seeds and therefore require no external data. The real-data benchmarks used in the experiments are publicly available from the sources described below:

\begin{itemize}[leftmargin=*]

  \item \textbf{Polyp segmentation:}
  the prepared polyp-segmentation dataset, which pools images from Kvasir-SEG \citep{jha2020kvasir}, CVC-ClinicDB \citep{bernal2015wmdova}, CVC-ColonDB \citep{tajbakhsh2016automated}, ETIS-LaribPolypDB \citep{silva2014toward}, and CVC-300 \citep{vazquez2017benchmark}, is obtained from the official AA-CRC repository, which provides separate \href{https://drive.google.com/file/d/1Y2z7FD5p5y31vkZwQQomXFRB0HutHyao/view}{training}
  and \href{https://drive.google.com/file/d/1YiGHLw4iTvKdvbT6MgwO9zcCv8zJ_Bnb/view}{test} archives. Pixelwise prediction scores are produced with the PraNet segmentation model \citep{fan2020pranet}.

  \item \textbf{RCV1 text multilabel classification:}
  the RCV1-v2 feature vectors and topic labels  \citep{lewis2004rcv1} are obtained using the \texttt{fetch\_rcv1} loader provided by
  \href{https://scikit-learn.org/stable/modules/generated/sklearn.datasets.fetch_rcv1.html}{scikit-learn}.

  \item \textbf{Letter Recognition:}
  the dataset  \citep{frey1991letter} is obtained from the
  \href{https://doi.org/10.24432/C5ZP40}{UCI Machine Learning Repository}.

  \item \textbf{Medical Insurance:}
  the dataset  \citep{lantz2013machine} is obtained from the \emph{Machine Learning with R} datasets
  \href{https://github.com/stedy/Machine-Learning-with-R-datasets/blob/master/insurance.csv}{repository on GitHub}.
  
  \item \textbf{Superconductor:}
  the dataset  \citep{hamidieh2018data} is obtained from the
  \href{https://doi.org/10.24432/C53P47}{UCI Machine Learning Repository}.
\end{itemize}

\bibliography{bibliography.bib}

\phantomsection\label{supplementary-material}
\bigskip

\appendix

\section{Algorithms}
\label{app:algorithms}

\begin{algorithm}[H]
\caption{\recirc{}: rectified conformal risk control}
\label{alg:recirc}
\begin{algorithmic}[1]
\Require ordered family $\{f_\lambda:\lambda\in\Lam\}$; threshold grid
$\Lam_M$; loss $\loss\in[0,B]$; level $\alpha$; risk-training set $\cD$;
calibration set
$\cC=\{(X_j,Y_j)\}_{j=1}^n$; budget grid
$\mathcal A=\{0=a_0<a_1<\cdots<a_L\}$ in $[0,B]$
\State Train and tune $\Rhat_\cD(\lambda\mid x)$ using only $\cD$
(Algorithm~\ref{alg:route-r}; see Appendix~\ref{app:response-model} for the
response-model alternative)
\For{$\ell=0,\ldots,L$}
  \For{$j=1,\ldots,n$}
    \State $\lambda_{j\ell}\gets \Rhat_\cD^{-1}(a_\ell\mid X_j)$
      \Comment{grid inversion of \eqref{eq:est-inverse}}
    \State $L_{j\ell}\gets \loss(f_{\lambda_{j\ell}}(X_j),Y_j)$
  \EndFor
  \State $\widehat{\mathcal R}^R_{\rm cal}(a_\ell)\gets
    n^{-1}\sum_{j=1}^n L_{j\ell}$
\EndFor
\State $\widehat a\gets\max\left\{a_\ell:
  \tfrac{n}{n+1}\widehat{\mathcal R}^R_{\rm cal}(a_\ell)+\tfrac{B}{n+1}
  \le\alpha\right\}$
\State \Return predictor $x\mapsto f_{\Rhat_\cD^{-1}(\widehat a\mid x)}(x)$
\end{algorithmic}
\end{algorithm}

\begin{algorithm}[H]
\caption{Direct risk-curve estimation via loss regression}
\label{alg:route-r}
\begin{algorithmic}[1]
\Require risk-training set $\cD=\{(X_i,Y_i)\}_{i=1}^{m}$; design distribution $q$
on $\Lam$; number of augmentations $K$; regression learner $g_\theta$;
threshold grid $\Lam_M$
\For{$i=1,\ldots,m$ and $k=1,\ldots,K$}
  \State draw $\lambda_{ik}\sim q$;\quad
    $Z_{ik}\gets\loss(f_{\lambda_{ik}}(X_i),Y_i)$
\EndFor
\State Train $g_\theta(x,\lambda)$ on
$\{(X_i,\lambda_{ik},Z_{ik})\}$ to approximate \eqref{eq:route-r}
\State For each queried $x$, let $\{\Rhat_\cD(\lambda\mid x)\}_{\lambda\in\Lam_M}$ be the
non-increasing (isotonic) regression of $\{g_\theta(x,\lambda)\}_{\lambda\in\Lam_M}$ on $\lambda$
\State \Return 
the monotonized curves
$\Rhat_\cD(\cdot\mid x)$ on $\Lam_M$
\end{algorithmic}
\end{algorithm}

\section{Response-model risk-curve estimation}
\label{app:response-model}

When a model for $Y\mid x$ is fitted on the risk-training set $\cD$ and
supplies a predictive distribution $\widehat p(y\mid x)$ from which one can
draw samples, the risk curve can
instead be estimated by modeling $Y\mid x$ and averaging the loss. On the
threshold grid $\Lam_M$, for every calibration or test input $x$, draw Monte
Carlo responses
$Y^{(1)}(x),\ldots,Y^{(S)}(x)\sim\widehat p(\cdot\mid x)$ and set
\begin{equation*}
  \Rhat_{\cD}(\lambda\mid x)
  =
  \frac1S\sum_{s=1}^S
  \loss(f_{\lambda}(x),Y^{(s)}(x)),
  \qquad \lambda\in\Lam_M.
\end{equation*}
If $\lambda\mapsto\loss(f_\lambda(x),y)$ is monotone for every $y$, using
the same sampled responses for all $\lambda\in\Lam_M$ makes the Monte Carlo curve
automatically monotone. The estimated curve is then inverted on the grid as
in \eqref{eq:est-inverse}. Algorithm \ref{alg:route-y} summarizes the procedure.

This alternative is useful when the base model already supplies reliable
samples or probabilities for $Y\mid x$: the same samples can be reused across
losses and decision families, and pointwise monotonicity is preserved without
post-processing. Its disadvantages are the cost of sampling or scoring a large
output space and the need to model aspects of the conditional law that may be
irrelevant to the loss of interest. Direct loss regression avoids those costs
by targeting a scalar response, at the price of a loss-specific fit and
explicit monotonicity handling.

\begin{algorithm}[H]
\caption{Alternative risk-curve estimation via a model for $Y\mid x$}
\label{alg:route-y}
\begin{algorithmic}[1]
\Require predictive model $\widehat p(y\mid x)$ fitted on $\cD$; threshold
grid $\Lam_M$; Monte Carlo size $S$; query point $x$
\State Draw $Y^{(1)}(x),\ldots,Y^{(S)}(x)\sim\widehat p(\cdot\mid x)$
\For{$\lambda\in\Lam_M$}
  \State $\Rhat_\cD(\lambda\mid x)\gets
    S^{-1}\sum_{s=1}^S \loss(f_{\lambda}(x),Y^{(s)}(x))$
    \Comment{same draws for all $\lambda$ preserve monotonicity}
\EndFor
\State \Return $\{\Rhat_\cD(\lambda\mid x)\}_{\lambda\in\Lam_M}$
\end{algorithmic}
\end{algorithm}

\section{Experimental settings and implementation details}
\label{app:experimental-settings}

\begin{table}[H]
\centering
\singlespacing
\caption{Overview of the experimental settings. All experiments are repeated 20 times. The AA-CRC feature map column summarizes how the input-dependent score threshold is parameterized. In the RF-based settings, a random forest is fitted on $D$ using a task-specific scalar measure of prediction difficulty, and $\Phi(x)$ is the concatenation of one-hot indicators of the leaves reached by $x$. Split sizes refer to the risk-training partition $\cD$, calibration partition $\cC$, and test partition $T$ of \recirc{}. Base-model fitting, \recirc{} grids and representations, AA-CRC regularization, and evaluation-group definitions are given in the subsections below.}
\footnotesize
\setlength{\tabcolsep}{4pt}
\begin{tabular}{@{}>{\raggedright\arraybackslash}p{2.5cm}
>{\raggedright\arraybackslash}p{4.9cm}
>{\raggedright\arraybackslash}p{3.1cm}
>{\centering\arraybackslash}p{4.5cm}
@{}}
\toprule
Setting & Base model & AA-CRC feature map & Splits (risk train / cal / test) \\
\midrule
\multicolumn{4}{@{}l}{\emph{Synthetic settings}} \\
\addlinespace[2pt]
Heteroscedastic & GAM mean with absolute-residual intervals (regression)
  & RF leaf indicators; RF target: absolute residual
& 1{,}000 / 500 / 1{,}000 \\
\addlinespace[4pt]
Latent difficulty & Synthetic label scores, 50 labels (multilabel)
   & Linear: intercept, difficulty covariate and score summaries (standardized on $D$)
  & 1{,}000 / 500 / 1{,}000 \\
\addlinespace[4pt]
XOR interaction & Quantile random forest (regression)
  & RF leaf indicators; RF target: minimum covering multiplier
  & 6{,}400 / 4{,}800 / 4{,}800 \\
\midrule
\multicolumn{4}{@{}l}{\emph{Real benchmarks}} \\
\addlinespace[2pt]
Polyp segmentation & PraNet pixelwise scores (segmentation)
  & 90 probability quantiles (label-free)
  & 700 / 700 / 398 \\
\addlinespace[4pt]
RCV1 & One-vs-rest logistic on TF--IDF, 103 labels (multilabel)
  & Score summaries and all 103 label scores
  & 2{,}000 / 1{,}500 / 14{,}500 \\
\addlinespace[4pt]
Letter Recognition & Logistic classifier, 26 classes (multiclass)
  & Input covariates and probability summaries
  & 6{,}000 / 4{,}000 / 4{,}000 \\
\addlinespace[4pt]
Medical Insurance & Quantile random forest (regression)
  & RF leaf indicators; RF target: minimum covering multiplier
  & 535 / 401 / 402 \\
\addlinespace[4pt]
Superconductor & Quantile random forest (regression)
  & RF leaf indicators; RF target: minimum covering multiplier
  & 8{,}505 / 6{,}378 / 6{,}380 \\
\bottomrule
\end{tabular}
\end{table}

\subsection{Base-model fitting and risk-training predictions}

The base predictive models and the samples used to fit them are as follows. In the heteroscedastic experiment, the GAM mean model is fitted on an independent sample of size $500$, separate from the $1{,}000$ observations in the risk-training set $\cD$. In the synthetic multilabel experiment, the label scores are generated directly from the data-generating mechanism, so there is no separately fitted base predictor. In polyp segmentation, we use the precomputed PraNet probability maps supplied with the benchmark and do not retrain PraNet.

For RCV1, the one-vs-rest logistic classifier is fitted on a separate base sample of $12{,}000$ documents, after which its predicted probabilities are computed on the conformal pool used for risk training, calibration, and evaluation. For Letter Recognition, the logistic classifier is fitted on the
30\% base split, corresponding to $6{,}000$ observations; the remaining data are split into risk-training, calibration, and test samples.

Medical Insurance, Superconductor, and XOR use quantile random forests (QRFs). In these three settings, the QRF is fitted on $\cD$. For observations in $\cD$, however, the QRF-derived location and scale features and the losses used to
train the \recirc{} risk estimator are computed from out-of-bag quantile predictions. Thus, an observation's own QRF fit is not used to construct its risk-regression target. For calibration and test observations, ordinary QRF predictions from the model fitted on $\cD$ are used.

\subsection{\recirc{} implementation details}
\label{app:recirc-implementation}

The heteroscedastic experiment uses 101 equally spaced values of $\lambda$ in $[0,4]$, 101 risk budgets in $[0,1]$, and $K=15$ threshold draws from a uniform distribution on $[0,4]$ per risk-training observation. The latent-difficulty experiment uses 101-point threshold and budget grids on $[0,1]$, with $K=20$ thresholds sampled uniformly from the threshold grid. Polyp segmentation uses 51 thresholds on $[0,1]$, 101 budget values on $[0,1]$, and $K=20$ uniform threshold draws on $[0,1]$. RCV1 uses all 26 equally spaced threshold values in $[0,1]$ for every risk-training observation and a 101-point budget grid on $[0,1]$. Letter Recognition uses the top-$m$ family, $m=1,\ldots,26$, with $K=8$ sampled values of $m$ and an 81-point budget grid on $[0,1]$. Medical Insurance, Superconductor, and XOR use 80 thresholds on $[0,4]$, 201 budgets on $[0,1]$, and 16 deterministic threshold anchors.

The nested family is the symmetric interval $[\widehat\mu(x)-\lambda,\widehat\mu(x)+\lambda]$ in the heteroscedastic experiment. The synthetic multilabel experiment includes labels with scores at least $1-\lambda$, and the polyp experiment includes pixels whose PraNet probabilities are at least $1-\lambda$. RCV1 uses the score-threshold family internally; equivalently, writing $\lambda=1-\tau$ gives the protective parameterization $\{k:\widehat p_k(x)\ge 1-\lambda\}$. Letter Recognition uses the $m$ classes with largest predicted probabilities.

In Medical Insurance, Superconductor, and XOR, the interval is centered at the estimated conditional median. Its lower and upper scales are given by the differences between the estimated $0.50$ and $0.05$ quantiles and between the estimated $0.95$ and $0.50$ quantiles, respectively, with both scales multiplied by $\lambda$. These experiments use the asymmetric miscoverage loss defined in Section~\ref{sec:exp-metrics}.

Medical Insurance, Superconductor, and XOR use the anchor implementation described in Section~\ref{sec:estimation}. Separate TabICLv2 regressors are fitted at 16 approximately equally spaced threshold anchors. The predicted anchor risks are clipped and made non-increasing in $\lambda$, interpolated to the complete 80-point grid using PCHIP, and monotonized once more before inversion. 

In the regression settings, the numerical threshold grid ends at $\lambda=4$, which need not be fully protective for every possible response. The theoretical family may be viewed as augmented by a fully protective endpoint, corresponding to the whole response space, which serves only as a fallback ensuring feasibility at $a_0=0$. This fallback was not required in any of the reported repetitions: the calibrated budget satisfied $\widehat a>0$ throughout, so all reported predictions were obtained from the finite numerical grid.

\subsection{Risk-regression representations}
\label{app:risk-representations}

The representation supplied to the conditional-risk estimator is task-specific. The heteroscedastic experiment uses $x$ together with $\lambda$. The synthetic multilabel experiment uses the observed covariates together with the mean, standard deviation, maximum, mean of the five largest scores, sum, and gap between the two largest label scores. Polyp segmentation represents each PraNet probability map by 90 empirical quantiles. RCV1 uses the three largest label scores, their sum, mean and standard deviation, summed binary entropy, counts of scores above $0.05$, $0.10$, $0.20$, and $0.50$, together with all 103 individual scores. Letter Recognition combines the original covariates with the five largest predicted probabilities, normalized predictive entropy, the top-two probability gap, top-three and top-five probability mass, the proportions of classes with probability at least $0.05$ and $0.10$, and the standard deviation of the probability vector. The corresponding decision coordinate is appended in the joint risk-regression settings.

For Medical Insurance, Superconductor, and XOR, the representation contains the original covariates, the estimated conditional median, the lower and upper QRF scales, their sum, the relative contribution of the upper scale, and the logarithm of the total scale. For observations in $\cD$, these QRF-derived quantities and the corresponding loss targets are computed from out-of-bag predictions.

TabICLv2 is used throughout. RCV1 uses four estimators with a context of $2{,}000$ risk-training observations. Letter Recognition limits the augmented risk-training sample to $6{,}000$ rows. The QRF-based experiments use four TabICLv2 estimators per anchor; Superconductor and XOR limit the risk-training context to $4{,}000$ observations per repetition.

\subsection{AA-CRC implementation}
\label{app:aacrc-implementation}

All AA-CRC variants use the objective and gradient of the authors' implementation and are optimized with SLSQP. The RF-leaf variants (heteroscedastic, Medical Insurance, Superconductor, and XOR) use no ridge penalty and impose box constraints on the coefficients; in the QRF-based settings, the minimum-covering-multiplier target used to train the random forest on $\cD$ is computed from out-of-bag QRF predictions. The latent-difficulty, RCV1, and Letter Recognition variants use ridge coefficient $0.01$ on the slope coefficients only, and polyp segmentation uses it on all coefficients; in Letter Recognition, the coefficient is increased by a factor of 10 if the optimizer reaches the numerical boundary.

As in the original procedure, $\theta$ is fitted on $\cC$ and deployed without further calibration, except in two settings. In polyp segmentation, $\theta$ is fitted on $\cD\cup\cC$, which, if anything, favors AA-CRC; its objective is evaluated on $64\times64$ downsampled masks, excluding empty ones, with the same 90 probability quantiles used by \recirc{} as features. In Letter Recognition, the original configuration gave a marginal risk of about $0.05$ at $\alpha=0.10$, because rank-based scores confine the attainable risk to the steps of the top-$m$ family; we therefore use the continuous adaptive-prediction-set score \citep{romano2020classification}, fit $\theta$ on $\cD$, and calibrate a scalar offset on $\cC$ by CRC, mirroring the data allocation of \recirc{} and restoring the marginal guarantee of AA-CRC. In all settings except Letter Recognition, because $\theta$ is fitted on $\cC$ and deployed without the test point in the optimization, the marginal guarantee of AA-CRC holds only approximately, with a slack that grows with the dimension of $\Phi$; this accounts for its marginal risk slightly above $\alpha$ in some settings, notably RCV1.

\subsection{Evaluation groups}
\label{app:evaluation-groups}

The groups used for worst-group risk and mean positive group excess are evaluation devices only and are not supplied to any method. The heteroscedastic experiment uses five equal-mass bins of the true conditional scale $\sigma(x)=0.2+0.6|x|$, while the latent-difficulty experiment uses five equal-mass bins of the known latent difficulty. Polyp segmentation uses five bins of a label-free uncertainty score based on the average proximity of the PraNet probabilities to $0.5$, with cutpoints fitted on $\cD$. RCV1 uses ten bins of summed binary predictive entropy, also with cutpoints fitted on $D$, while Letter Recognition uses five equal-mass bins of normalized predictive entropy.

Medical Insurance uses eight potentially overlapping groups obtained by crossing smoker status with the indicator $\mathrm{BMI}\ge 30$ and, separately, with an indicator for age above its sample median; groups with fewer than 30 test observations are omitted. Superconductor uses the four intersections obtained by median-splitting \texttt{range\_ThermalConductivity} and \texttt{wtd\_std\_ThermalConductivity}; groups with fewer than 50 test observations are omitted. Because these two variables were selected using their association with the response in the full dataset, this grouped analysis is exploratory. XOR uses the four sign quadrants of $(X_1,X_2)$.

\section{\recirc{} Calibration Diagnostics}
\label{app:calibration-diagnostics}

Each figure includes a panel for the overall group $G=\mathcal X$. The additional groups were fixed within each repetition as follows. In the heteroscedastic and latent-difficulty simulations, they are five equal-mass bins of, respectively,
the known noise scale $\sigma(X)$ and the known latent difficulty. For polyp segmentation, they are five bins of the label-free score $|\mathcal U|^{-1}\sum_{u\in\mathcal U}\{1-2|\widehat p_u(X)-1/2|\}$, with cutpoints fitted on the
risk-training images. For RCV1, they are ten bins of the summed binary entropy of the 103 label scores, with cutpoints fitted on the risk-training split; for Letter Recognition, they are five equal-mass bins of normalized predictive
entropy; and for Medical Insurance, five equal-mass bins of the total QRF scale $s_-(X)+s_+(X)$. The Superconductor groups are the four intersections obtained by median-splitting the two covariates with the largest absolute full-data correlations with the response (\texttt{range\_ThermalConductivity} and \texttt{wtd\_std\_ThermalConductivity}); because this selection used the responses, that diagnostic is exploratory. Finally, the XOR groups are the four sign quadrants of $(X_1,X_2)$. Except for the stated Superconductor
exception, group construction uses only covariates, base-model outputs, or known simulation quantities, and never the held-out outcomes used to compute $\widehat r_G(a)$.

\begin{figure}[H]
    \centering
    \includegraphics[width=\linewidth]{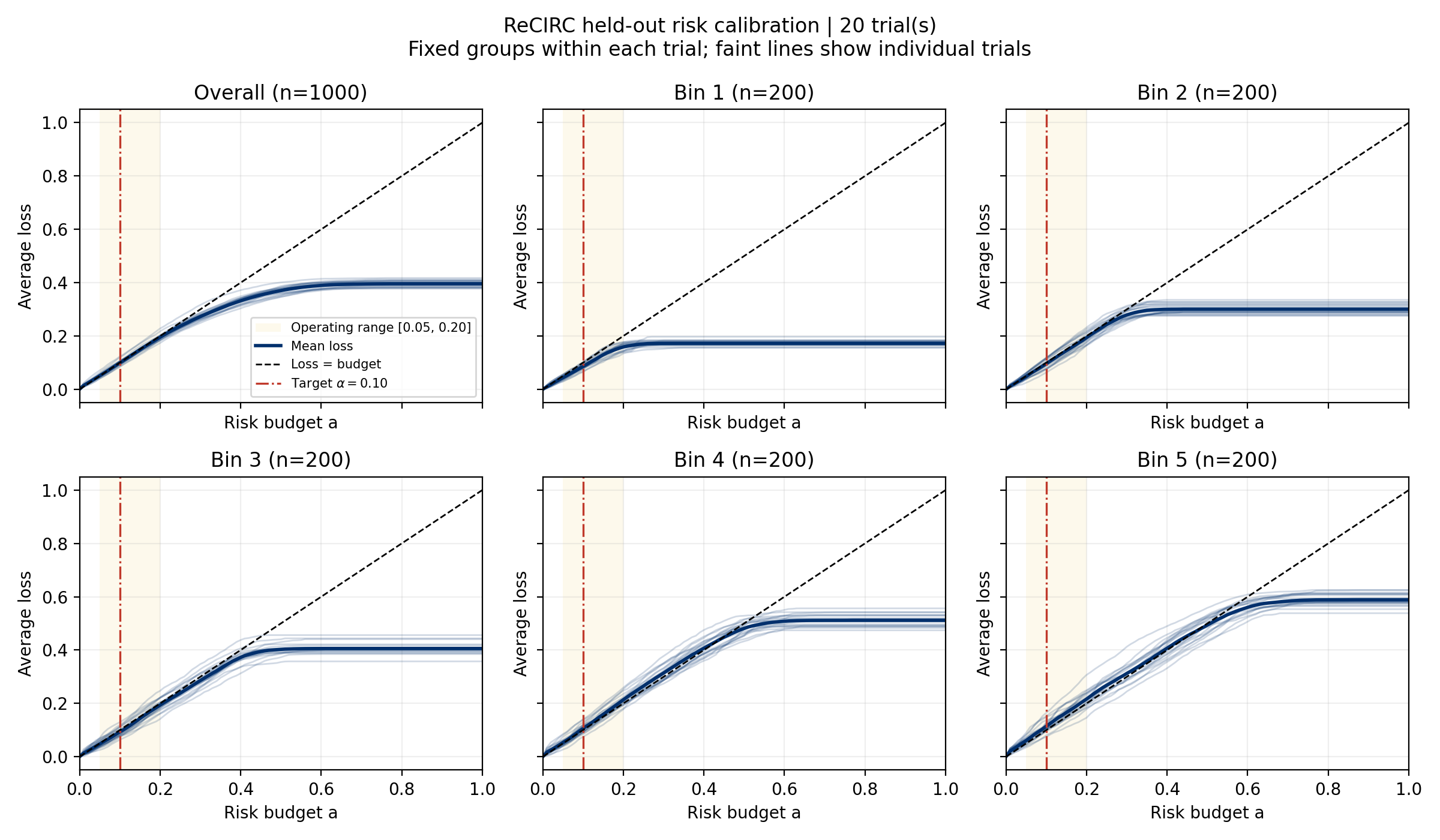}
    \caption{Held-out risk-calibration diagnostics for \recirc{} in the heteroscedastic experiment, overall and within five equal-mass bins of the known noise scale $\sigma(X)$. Thick curves show mean empirical loss over 20 repetitions; faint curves show individual repetitions, with groups fixed
    within each repetition.}
    \label{fig:risk-calibration-1}
\end{figure}

\begin{figure}[H]
    \centering
    \includegraphics[width=\linewidth]{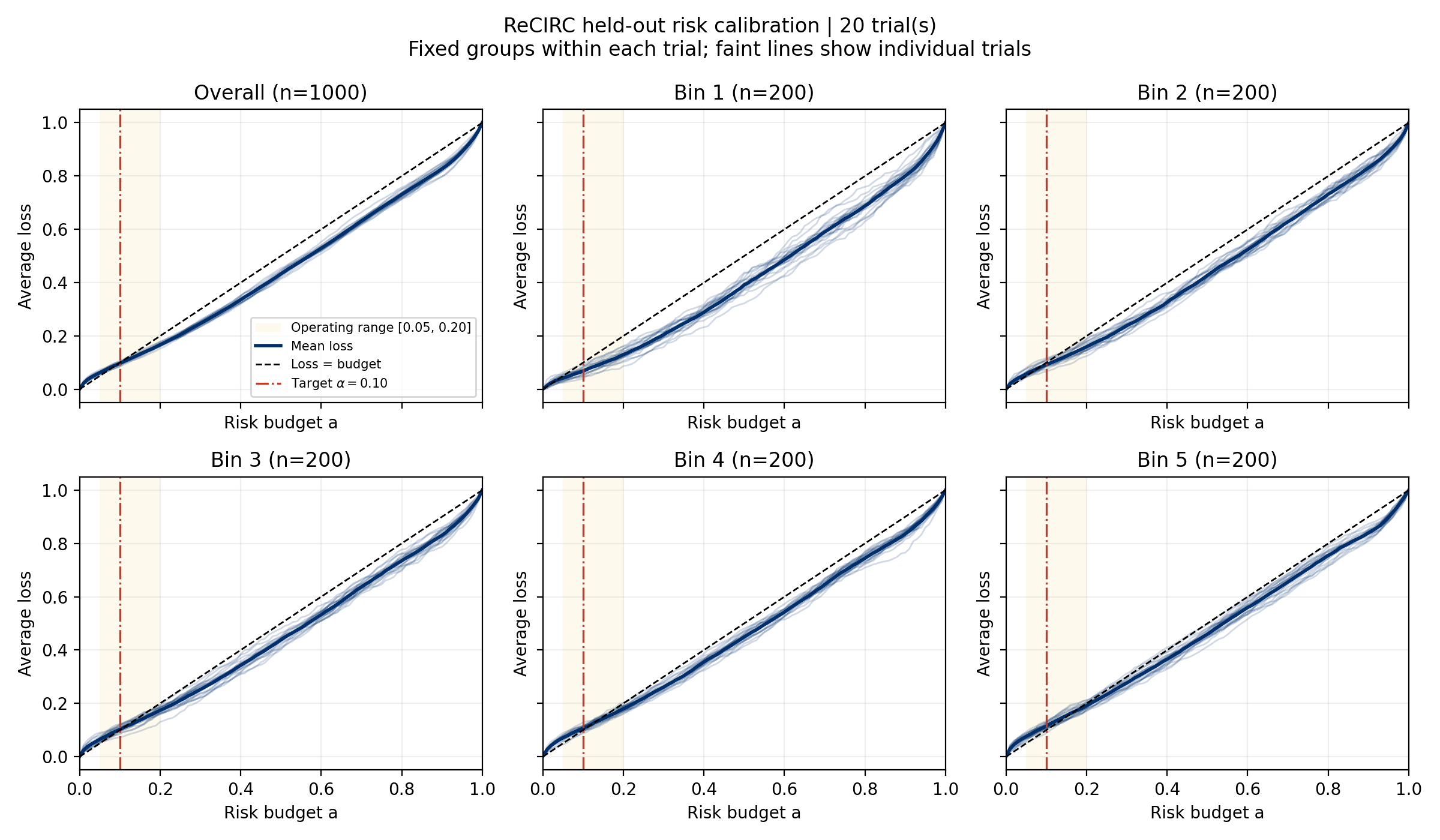}
    \caption{Held-out risk-calibration diagnostics for \recirc{} in the latent-difficulty experiment, overall and within five equal-mass bins of the known latent difficulty. Thick curves show mean empirical loss over 20 repetitions;
    faint curves show individual repetitions, with groups fixed
    within each repetition.}
    \label{fig:risk-calibration-2}
\end{figure}

\begin{figure}[H]
    \centering
    \includegraphics[width=\linewidth]{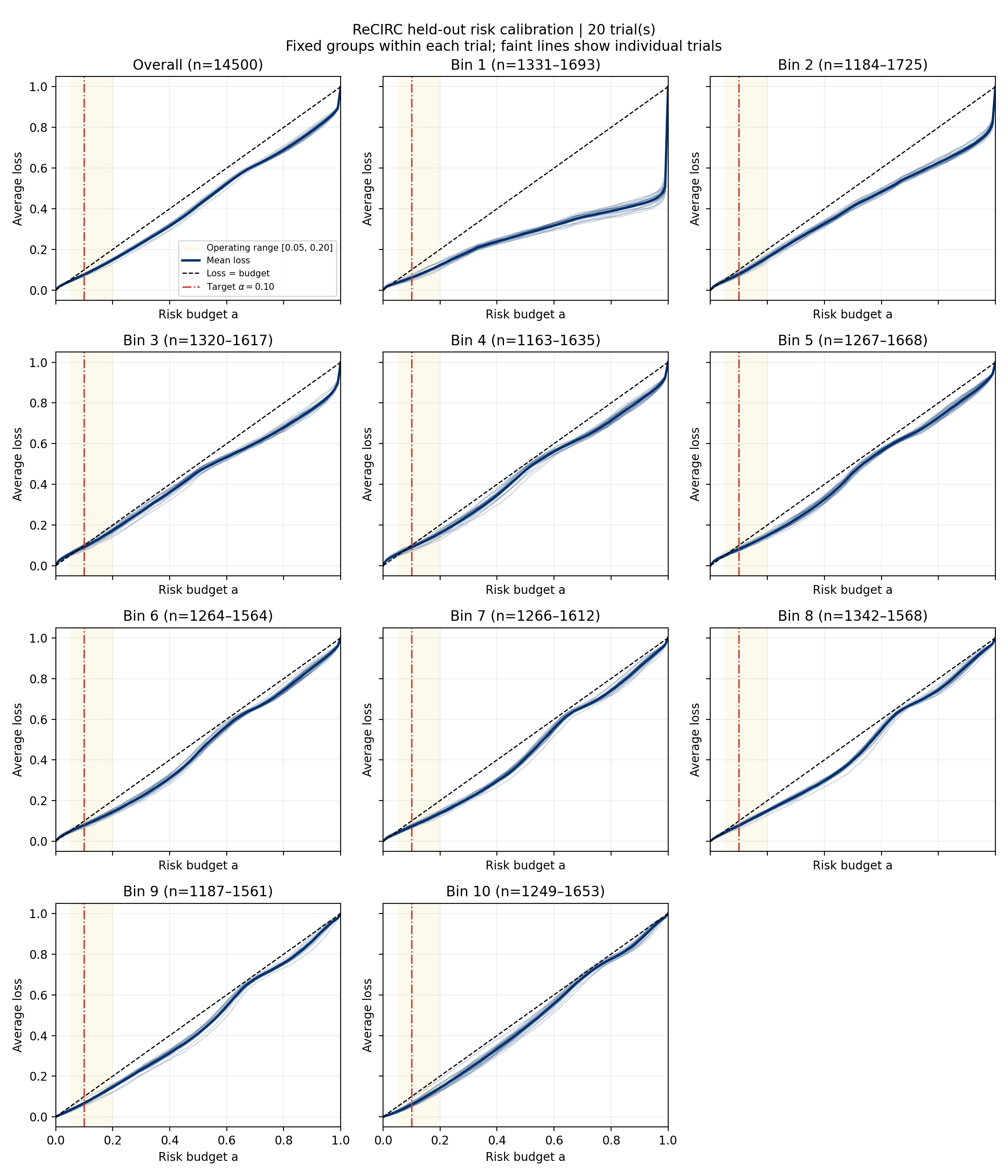}
    \caption{Held-out risk-calibration diagnostics for \recirc{} in RCV1. Results are shown over 20 repetitions, using fixed groups within each repetition; faint curves correspond to individual repetitions.}
    \label{fig:risk-calibration-4}
\end{figure}

\begin{figure}[H]
    \centering
    \includegraphics[width=\linewidth]{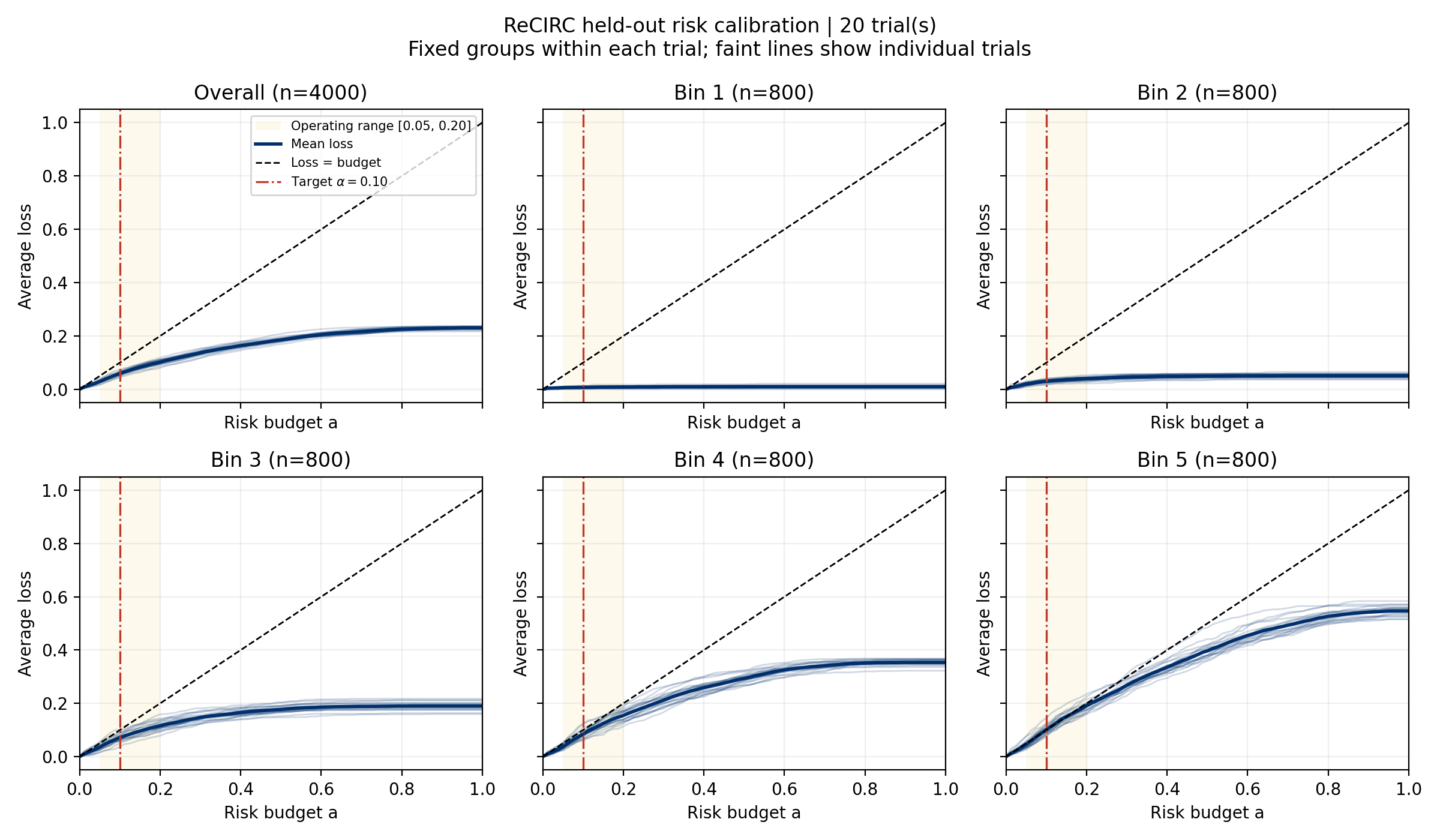}
    \caption{Held-out risk-calibration diagnostics for \recirc{} in Letter Recognition. Results are shown over 20 repetitions, using fixed groups within each repetition; faint curves correspond to individual repetitions.}
    \label{fig:risk-calibration-5}
\end{figure}

\begin{figure}[H]
    \centering
    \includegraphics[width=\linewidth]{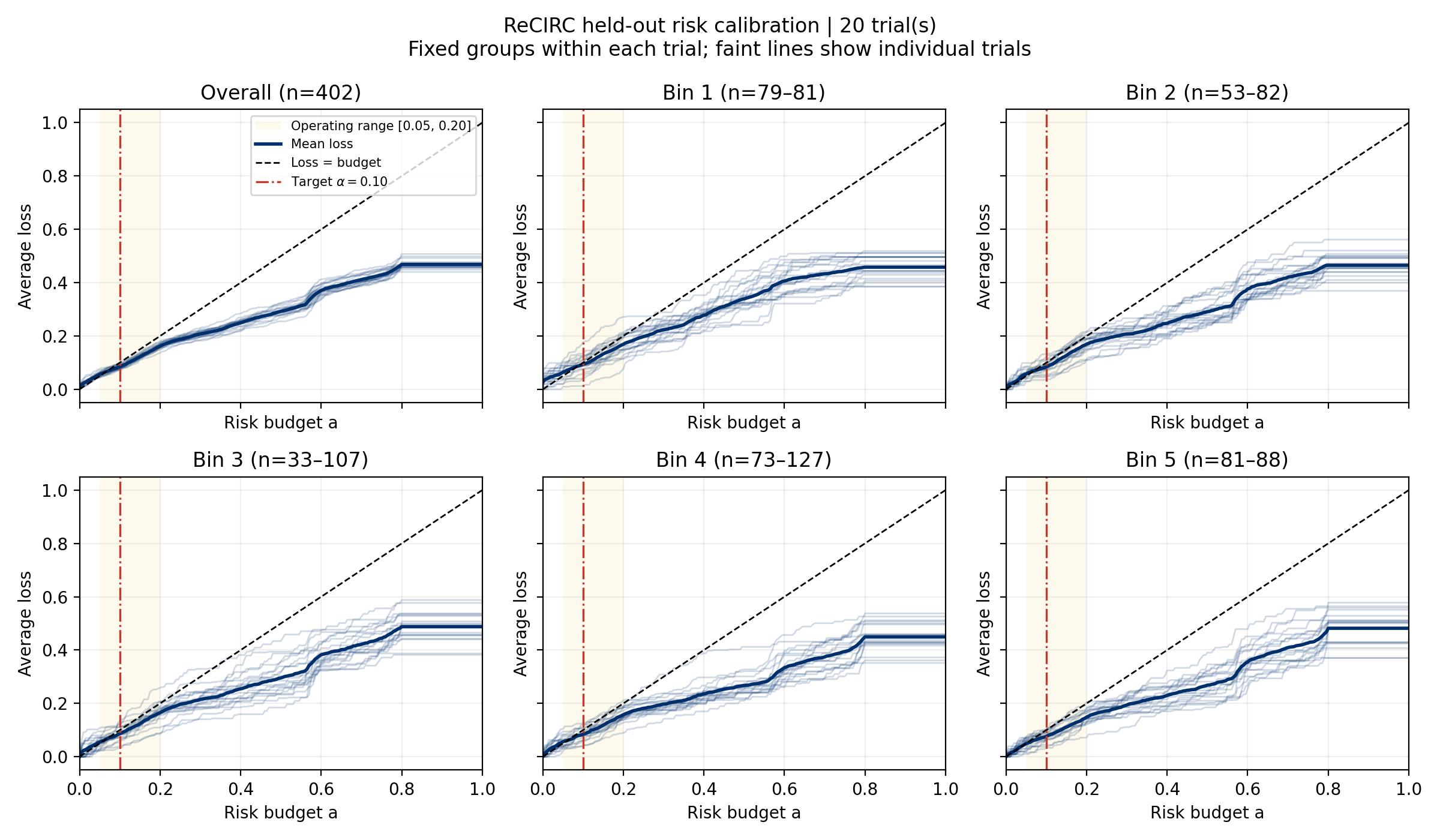}
    \caption{Held-out risk-calibration diagnostics for \recirc{} in Medical Insurance. Results are shown over 20 repetitions, using fixed groups within each repetition; faint curves correspond to individual repetitions.}
    \label{fig:risk-calibration-6}
\end{figure}

\begin{figure}[H]
    \centering
    \includegraphics[width=\linewidth]{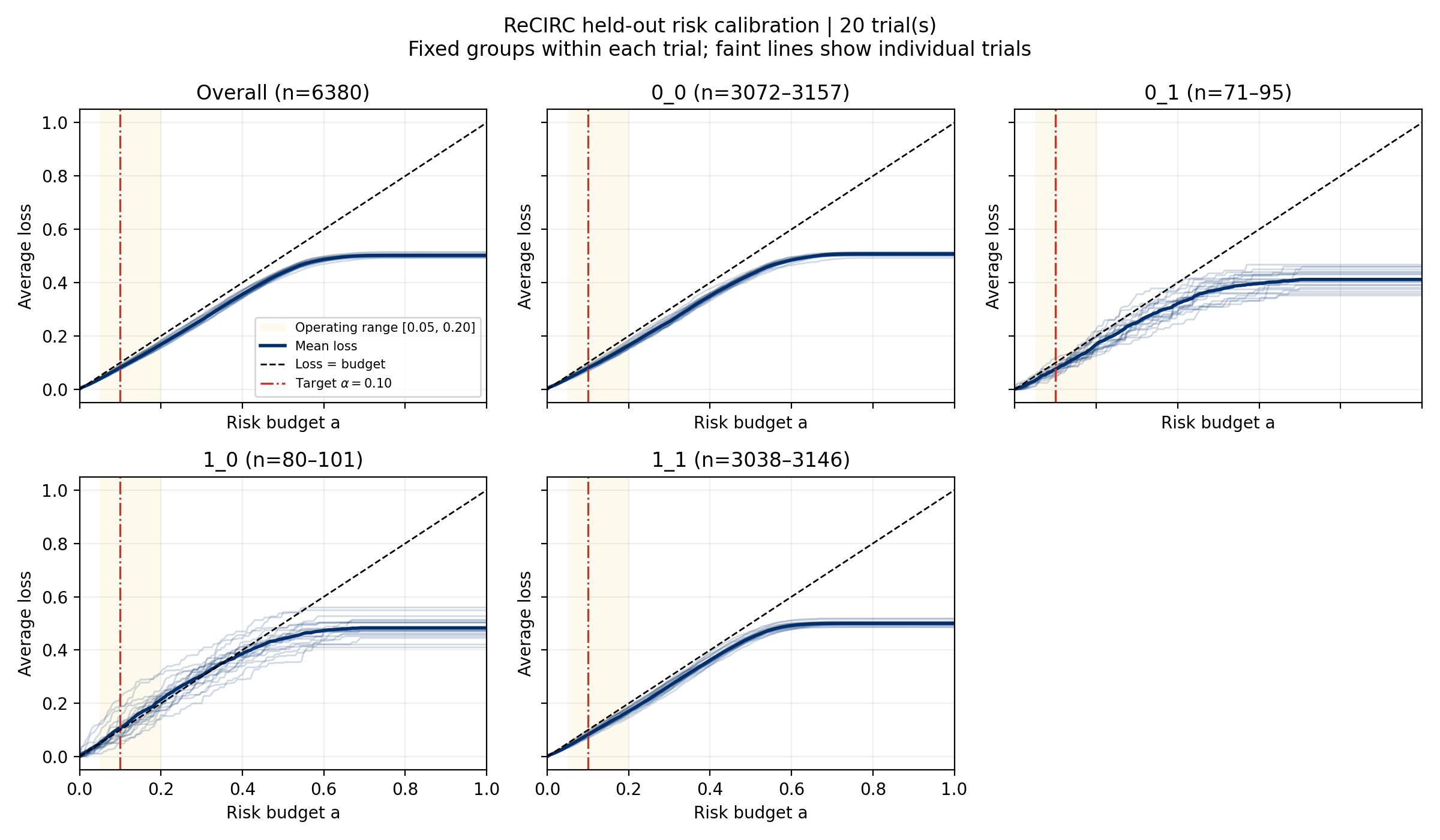}
    \caption{Held-out risk-calibration diagnostics for \recirc{} in Superconductor. Results are shown over 20 repetitions, using fixed groups within each repetition; faint curves correspond to individual repetitions.}
    \label{fig:risk-calibration-7}
\end{figure}

\begin{figure}[H]
    \centering
    \includegraphics[width=\linewidth]{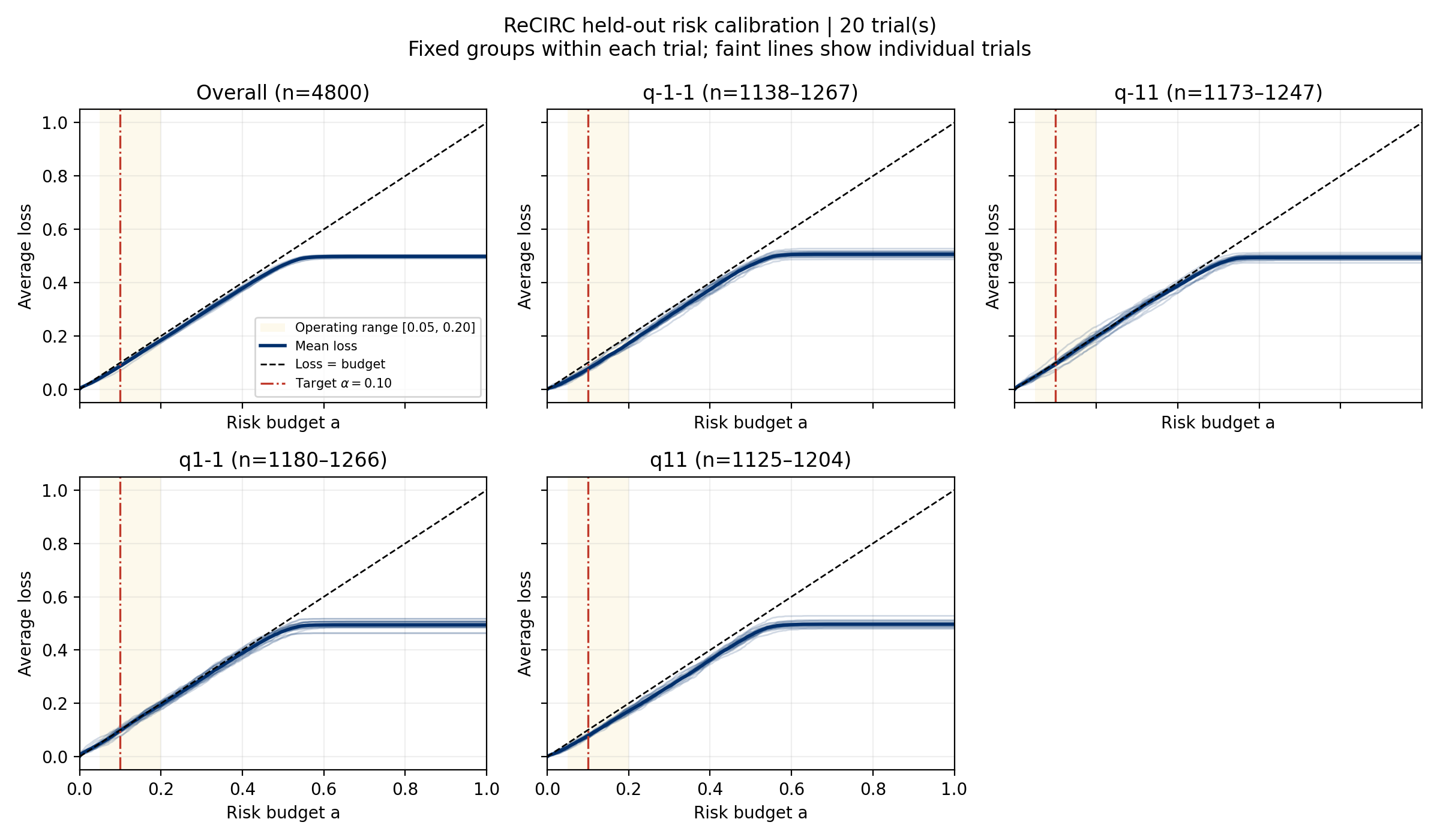}
    \caption{Held-out risk-calibration diagnostics for \recirc{} in the XOR interaction experiment. Results are shown over 20 repetitions, using fixed groups within each repetition; faint curves correspond to individual repetitions.}
    \label{fig:risk-calibration-8}
\end{figure}

Figure~\ref{fig:risk-calibration-3} in Section~\ref{sec:experiments} and the seven figures in this appendix together show the held-out risk-calibration diagnostics for \recirc{} across the eight experimental settings. Each figure plots empirical average loss $\widehat r_G(a)$ against the risk budget $a$. Thick blue curves
show means over 20 repetitions; faint curves show individual repetitions, with groups fixed within each repetition. The dashed black diagonal marks equality between risk and budget, the red dash-dotted vertical line marks $\alpha=0.10$, and the shaded band highlights $a\in[0.05,0.20]$. Curves below the diagonal indicate
conservative risk relative to the advertised budget. Such departures can arise from risk-estimation error, discrete decision choices, or saturation at the least protective endpoint.  In the regression settings, the plateaus arise because the interval at $\lambda=0$ collapses to a point whose risk caps the attainable budget: about $0.5$ under $\loss_{\mathrm{asym}}$, and increasing with the noise scale under $\loss_{\mathrm{exc}}$. In Letter Recognition, \recirc{} selects among nonempty top-$m$ prediction sets. Once $m=1$, further increases in the budget leave the
prediction unchanged, and the loss equals the top-1 classification error indicator. This explains the low plateaus in the low-entropy groups. Moreover, the discrete choices of $m$ can produce conservative risk even before saturation and even with exact conditional-risk estimates.  Departures above the diagonal are small, at most about $0.03$, and occur mainly in the noisiest heteroscedastic bins and, at small budgets, in the harder latent-difficulty bins.

\section{Additional theoretical results and representations}
\label{app:representations}

\subsection{Most-input and group consequences}
\label{app:aggregate-conditional}

\begin{proposition}[Most-input and group conditional-risk bounds]
\label{prop:aggregate-conditional}
Under the conditions of Proposition~\ref{prop:approx-conditional}, the
following conclusions hold conditionally on $\cD$, for almost every
realization of $\cD$; throughout, $\Prob_X$ and the $L^p(P_X)$ norm are
evaluated with that realization fixed. Suppose
$\varepsilon_M\in L^p(P_X)$ for some $p\in[1,\infty)$. Then, for every
$a\in[0,B]$ and $t>0$,
\begin{equation}
  \Prob_X\left\{x:
  \E\left[\loss\left(\widehat f^R_{\cD,a}(X),Y\right)
  \,\middle|\,X=x,\cD\right]>a+t
  \right\}
  \le
  \frac{\|\varepsilon_M\|_{L^p(P_X)}^p}{t^p}.
  \label{eq:most-input-bound}
\end{equation}
For every measurable group $G\subseteq\cX$ with
$\Prob_X(X\in G)>0$,
\begin{equation}
  \E\left\{
  \loss\left(\widehat f^R_{\cD,a}(X),Y\right)
  \,\middle|\,X\in G,\cD\right\}
  \le
  a+\Prob_X(X\in G)^{-1/p}\|\varepsilon_M\|_{L^p(P_X)}.
  \label{eq:group-conditional-bound}
\end{equation}
If the calibration and test pairs are i.i.d., the budget-grid condition holds,
$0<\alpha\le B$, and $n\ge B/\alpha-1$, both conclusions remain valid after conditioning on the
calibration set and replacing $a$ by $\widehat a$.
\end{proposition}

The group risk on the left side of \eqref{eq:group-conditional-bound} is
estimated directly by the held-out risk-calibration curve
\eqref{eq:risk-calibration-curve}. Thus the diagnostic provides a practical
check of the population quantity controlled by this proposition.

\subsection{Rate transfer from risk-curve estimation}

\begin{proposition}[Rate transfer from risk-curve estimation]
\label{prop:risk-rate}
Suppose Assumptions~\ref{ass:exchange} and~\ref{ass:monotone} hold, and let
$\cD_m$ be a risk-training sample. Let
$h_M:=\max_{j<M}(\lambda_{j+1}-\lambda_j)$. Suppose
$R(\cdot\mid x)$ is Lipschitz and, for a sequence $\eta_m\to0$, 
\[
  \max_{\lambda\in\Lam_M}
  |R(\lambda\mid x)-\Rhat_{\cD_m}(\lambda\mid x)|
  =O_{\Prob}(\eta_m).
\]
Then
\[
  \sup_{0\le a\le R(\lambda_{\min}\mid x)}
  \left|
  \E\left\{\loss\left(\widehat f^R_{\cD_m,a}(X),Y\right)
  \,\middle|\,X=x,\cD_m\right\}-a
  \right|
  =O_{\Prob}(\eta_m+h_M).
\]
If, in addition, the calibration and test pairs are i.i.d.,
Assumption~\ref{ass:regularity} holds, $0<\alpha\le B$,
$n\ge B/\alpha-1$, and, along the asymptotic sequence under consideration,
\[
  \Prob\{\widehat a\le R(\lambda_{\min}\mid x)\}\longrightarrow1,
\]
then also
\[
  \left|
  \E\left\{\loss\left(\widehat f^R_{\cD_m,\widehat a}(X),Y\right)
  \,\middle|\,X=x,\cD_m,\cC\right\}-\widehat a
  \right|
  =O_{\Prob}(\eta_m+h_M).
\]
\end{proposition}

\subsection{A general deployed-risk rate}

\begin{theorem}[General rate for the deployed conditional risk]
\label{thm:deployed-conditional}
Along a sequence $n\to\infty$, let the risk-training size satisfy
$m=m(n)\to\infty$ and let $M=M(n)$. All quantities below are indexed along
this sequence: $\rho_m$, $\eta_m$, $h_M$, and $h_n$ abbreviate
$\rho_{m(n)}$, $\eta_{m(n)}$, the spacing of $\Lam_{M(n)}$, and the spacing of
$\mathcal A_n$, and every $O_{\Prob}$ statement is taken as $n\to\infty$.
Suppose Assumptions~\ref{ass:exchange}--\ref{ass:regularity}
hold along this sequence, with the calibration and test pairs i.i.d. Let
$\cD_m$ be the risk-training sample, and let
$\mathcal A_n=\{0=a_{0,n}<\cdots<a_{L_n,n}=B\}$ have maximum spacing
$h_n\to0$. Fix $\alpha\in(0,B)$ and suppose $n\ge B/\alpha-1$.
For an open interval $I\subset(0,B)$ containing $\alpha$, suppose that, for
some deterministic $\rho_m\to0$,
\[
  \sup_{a\in I}
  \left|
  \E\left\{\loss\left(\widehat f^R_{\cD_m,a}(X),Y\right)
  \,\middle|\,\cD_m\right\}-a
  \right|
  =O_{\Prob}(\rho_m).
\]
Then
\[
  |\widehat a-\alpha|
  =O_{\Prob}(\rho_m+n^{-1/2}+h_n).
\]
If, in addition, for a fixed $x$, the interval $I$ lies in
$(0,R(\lambda_{\min}\mid x))$, $R(\cdot\mid x)$ is Lipschitz,
the threshold grid has maximum spacing
$h_M:=\max_{j<M}(\lambda_{j+1}-\lambda_j)\to0$, and, for a sequence
$\eta_m\to0$,
\[
  \max_{\lambda\in\Lam_M}
  |R(\lambda\mid x)-\Rhat_{\cD_m}(\lambda\mid x)|
  =O_{\Prob}(\eta_m),
\]
then
\[
  \left|
  \E\left\{\loss\left(\widehat f^R_{\cD_m,\widehat a}(X),Y\right)
  \,\middle|\,X=x,\cD_m,\cC\right\}-\alpha
  \right|
  =O_{\Prob}(\rho_m+\eta_m+h_M+n^{-1/2}+h_n).
\]
\end{theorem}

The $\rho_m$ condition is a high-level marginal risk-scale condition, not a
calibration conclusion. A primitive sufficient condition is uniform
risk-curve accuracy: if the error bound and Lipschitz constant in
Proposition~\ref{prop:risk-rate} hold uniformly over $x$, then that proposition
and integration over $X$ give $\rho_m$ of order $\eta_m+h_M$.

\subsection{Rectification as a local quantile transformation}
\label{sec:canonical}

The rectification map depends on the original family and loss only through the
local risk curve. This leads to a canonical representation that connects the
construction to an ordinary quantile transformation. Normalize $B=1$ for this
subsection.

\begin{proposition}[Canonical representation]
\label{prop:canonical}
Fix $x$ and assume $\lambda\mapsto R(\lambda\mid x)$ is non-increasing and
right-continuous, with $R(\lambda_{\max}\mid x)=0$. Then:
\begin{enumerate}[label=(\alph*)]
\item $F_x(\lambda):=1-R(\lambda\mid x)$ is a cumulative distribution
function, so there exists a random variable $Z_x$ with
$\Prob(Z_x>\lambda)=R(\lambda\mid x)$;
\item the risk-curve inverse is a generalized quantile:
\[
  R^{-1}(a\mid x)=F_x^{-1}(1-a),
  \qquad
  F_x^{-1}(u):=\inf\{\lambda\in\Lam:F_x(\lambda)\ge u\}.
\]
In particular, this convention gives $F_x^{-1}(0)=\lambda_{\min}$;
\item the oracle rectified rule satisfies
$f_a^R(x)=f_{F_x^{-1}(1-a)}(x)$, and the canonical problem defined by the
threshold rules $f^{\mathrm{can}}_\lambda(x)=\{z:z\le\lambda\}$ with $0$--$1$
loss $\ind\{z>\lambda\}$ applied to $Z_x$ has exactly the same local risk curve
and rectification map as the original problem.
\end{enumerate}
\end{proposition}

Thus risk rectification can be viewed as replacing the raw threshold by a
local quantile of a latent variable whose upper tail is the conditional risk
curve. In the miscoverage special case of
Section~\ref{sec:score-rectification}, this latent variable has the conditional
law of the score $S(x,Y)$; for a general bounded monotone loss, it is only a
representation of the risk curve and need not correspond to an observed
score.

\section{Proofs}
\label{app:proofs}

\subsection{Proof of Theorem \ref{thm:marginal-validity}}

Condition on $\cD$, so that the rectified family
$\{\widehat f^R_{\cD,a}:a\in[0,B]\}$ is fixed, and write
\[
  L_j(a)
  :=
  \loss\left(\widehat f^R_{\cD,a}(X_j),Y_j\right),
  \qquad j=1,\ldots,n+1 .
\]
We first record two structural facts. First, the map
$a\mapsto\Rhat^{-1}_\cD(a\mid x)$ is non-increasing on $[0,B]$: for $a>0$ the
sets $\{\lambda\in\Lam_M:\Rhat_\cD(\lambda\mid x)\le a\}$ grow with $a$,
so their minima decrease, and the convention
$\Rhat^{-1}_\cD(0\mid x)=\lambda_{\max}$
returns the largest possible threshold. By \eqref{eq:monotone-family}, each
$L_j$ is therefore nondecreasing in $a$, with values in $[0,B]$; no separate
regularity assumption is needed. Second, at $a=0$ the convention deploys
$f_{\lambda_{\max}}$, so $L_j(0)=0$ by Assumption~\ref{ass:monotone}, and the
condition $n\ge B/\alpha-1$ gives
$\tfrac{n}{n+1}\cdot 0+\tfrac{B}{n+1}\le\alpha$: the constraint set in
\eqref{eq:ahat} contains $a_0=0$, so $\widehat a$ is a well-defined maximum
over a finite nonempty set, and it satisfies the CRC constraint by
construction.

Now define the augmented selection rule
\[
  \widehat a'
  :=
  \max\left\{a\in\mathcal A:
  \frac{1}{n+1}\sum_{j=1}^{n+1}L_j(a)\le\alpha\right\},
\]
which is well defined for the same reason. If $a$ is feasible in
\eqref{eq:ahat}, then
$\frac{1}{n+1}\sum_{j=1}^{n+1}L_j(a)
\le\frac{n}{n+1}\widehat{\mathcal R}^R_{\rm cal}(a)+\frac{B}{n+1}\le\alpha$
because $L_{n+1}(a)\le B$; hence $\widehat a\le\widehat a'$, and monotonicity
of the losses gives $L_{n+1}(\widehat a)\le L_{n+1}(\widehat a')$.
Conditional on $\cD$, each $L_j(\cdot)$ is a fixed measurable transformation
of $(X_j,Y_j)$, so the loss paths $L_1,\ldots,L_{n+1}$ are exchangeable by
Assumption~\ref{ass:exchange}, and $\widehat a'$ is a symmetric function of
them. Therefore
\[
  \E\{L_{n+1}(\widehat a')\mid\cD\}
  =
  \E\left\{\frac{1}{n+1}\sum_{j=1}^{n+1}L_j(\widehat a')\,\middle|\,\cD\right\}
  \le\alpha,
\]
where the inequality holds pointwise because $\widehat a'$ is itself
feasible---this is where the finite grid matters, as it makes the maximum
attained. Combining the two displays gives
$\E\{L_{n+1}(\widehat a)\mid\cD\}\le\alpha$, which is the claimed inequality;
taking expectations over $\cD$ gives the unconditional statement. This is the
conformal risk control argument of \citet{angelopoulos2024conformal},
specialized to a finite ordered index set.

Finally, when the response-model alternative of
Appendix~\ref{app:response-model} draws fresh Monte Carlo responses at each
evaluated input, attach to each pair $(X_j,Y_j)$ an independent auxiliary
variable $\xi_j$ collecting its draws; the triples $(X_j,Y_j,\xi_j)$ remain
exchangeable, each $L_j(\cdot)$ is a fixed measurable transformation of its
triple, and the argument applies verbatim. \qed

\subsection{Proof of Propositions \ref{prop:oracle-conditional} and
\ref{prop:approx-conditional}}

For Proposition~\ref{prop:oracle-conditional}, write
$\lambda^\ast:=R^{-1}(a\mid x)=\inf\{\lambda:R(\lambda\mid x)\le a\}$.
Assumption~\ref{ass:monotone} gives
$R(\lambda_{\max}\mid x)=0\le a$, so the defining set is nonempty. Take
$\lambda_k\downarrow\lambda^\ast$ with
$R(\lambda_k\mid x)\le a$; right-continuity gives
$R(\lambda^\ast\mid x)=\lim_k R(\lambda_k\mid x)\le a$. The first display then
follows from \eqref{eq:oracle-bound}. For the equality claim, suppose
$R(\cdot\mid x)$ is continuous and $a\le R(\lambda_{\min}\mid x)$, and assume
for contradiction that $R(\lambda^\ast\mid x)<a$. Since
$R(\lambda_{\min}\mid x)\ge a$, we have $\lambda^\ast>\lambda_{\min}$, and by
continuity there exists $\lambda<\lambda^\ast$ with $R(\lambda\mid x)<a$,
contradicting the definition of $\lambda^\ast$ as the infimum. Hence
$R(\lambda^\ast\mid x)=a$.

For Proposition~\ref{prop:approx-conditional}, let
$\widehat\lambda_a(x):=\Rhat^{-1}_\cD(a\mid x)$. If $a=0$ or the defining
set is empty, then $\widehat\lambda_a(x)=\lambda_{\max}$ and the desired
bound is trivial by Assumption~\ref{ass:monotone}. Otherwise, the selected
threshold belongs to the finite feasible set, so
$\Rhat_\cD(\widehat\lambda_a(x)\mid x)\le a$. Therefore
\[
  \E\left\{\loss\left(\widehat f^R_{\cD,a}(X),Y\right)\,\middle|\, X=x,\cD\right\}
  =
  R(\widehat\lambda_a(x)\mid x)
  \le
  \Rhat_\cD(\widehat\lambda_a(x)\mid x)+\varepsilon_M(x)
  \le
  a+\varepsilon_M(x),
\]
where the equality uses that $\widehat\lambda_a(x)$ is $\cD$-measurable and
$(X,Y)$ is independent of $\cD$ (Assumption~\ref{ass:exchange}), and the first
inequality uses the gridwise one-sided error bound.
Under the additional i.i.d.\ condition, the test pair is independent of
$\cC$. Since the bound holds simultaneously in $a$, we may condition on
$\cC$ and substitute $a=\widehat a$. This proves
Proposition~\ref{prop:approx-conditional}. \qed

\subsection{Proof of Proposition \ref{prop:aggregate-conditional}}

Fix a realization of $\cD$ for which the pointwise conclusion of
Proposition~\ref{prop:approx-conditional} holds. Define
\[
  V_a(x)
  :=
  \left[
  \E\left\{\loss\left(\widehat f^R_{\cD,a}(X),Y\right)
  \,\middle|\,X=x,\cD\right\}-a
  \right]_+.
\]
The pointwise result gives $V_a(x)\le\varepsilon_M(x)$. Markov's inequality
applied to $V_a^p(X)$ proves \eqref{eq:most-input-bound}. For a measurable
group $G$ of probability $\pi_G>0$, H\"older's inequality yields
\[
  \E\{\varepsilon_M(X)\mid X\in G,\cD\}
  =
  \frac{\E\{\varepsilon_M(X)\ind(X\in G)\mid\cD\}}{\pi_G}
  \le
  \pi_G^{-1/p}\|\varepsilon_M\|_{L^p(P_X)}.
\]
Combining this with the pointwise result and averaging over $X\mid X\in G$
proves \eqref{eq:group-conditional-bound}. Under the additional i.i.d.\
condition, the test pair is independent of $\cC$. Since both bounds hold
simultaneously in $a$, we may condition on $\cC$ and substitute
$a=\widehat a$. \qed

\subsection{Proof of Proposition \ref{prop:risk-rate}}

Set $\Delta_m(x):=\max_{\lambda\in\Lam_M}
|R(\lambda\mid x)-\Rhat_{\cD_m}(\lambda\mid x)|$, and let $L_x$ be
a Lipschitz constant for $R(\cdot\mid x)$. Fix
$a\in[0,R(\lambda_{\min}\mid x)]$ and write
$\widehat\lambda_a(x)=\Rhat_{\cD_m}^{-1}(a\mid x)$. The proposition gives
$R(\widehat\lambda_a(x)\mid x)\le a+\Delta_m(x)$.

For the reverse inequality, first consider the boundary cases. If $a=0$,
the convention selects $\lambda_{\max}$ and the true risk is zero. If the
estimated feasible set is empty, then
$\Rhat_{\cD_m}(\lambda_{\max}\mid x)>a$ while
$R(\lambda_{\max}\mid x)=0$, so $a<\Delta_m(x)$. If
$\widehat\lambda_a(x)=\lambda_{\min}$, feasibility and
$a\le R(\lambda_{\min}\mid x)$ give
$0\le R(\lambda_{\min}\mid x)-a\le\Delta_m(x)$.
In every boundary case the claimed lower bound follows. Otherwise, let
$\lambda^-$ be the grid point immediately below
$\widehat\lambda_a(x)$. Its infeasibility gives
$R(\lambda^-\mid x)>a-\Delta_m(x)$, while Lipschitz continuity gives
$R(\widehat\lambda_a(x)\mid x)\ge R(\lambda^-\mid x)-L_xh_M$.
Thus the absolute error is at most $\Delta_m(x)+L_xh_M$, uniformly over all
attainable budgets. Under the additional conditions in the final sentence of
the proposition, the same bound applies at $a=\widehat a$ on the event
$\{\widehat a\le R(\lambda_{\min}\mid x)\}$. Since this event has probability
tending to one, the claimed $O_{\Prob}(\eta_m+h_M)$ conclusion follows. \qed

\subsection{Proof of Theorem \ref{thm:deployed-conditional}}

For compactness in this proof, write
\[
  R_m^R(a\mid x)
  :=
  \E\left\{\loss\left(\widehat f^R_{\cD_m,a}(X),Y\right)
  \,\middle|\,X=x,\cD_m\right\},
  \qquad
  r_m(a):=\E\{R_m^R(a\mid X)\mid\cD_m\},
\]
and let
\[
  q_m:=\sup_{a\in I}|r_m(a)-a|,
  \qquad
  q_m(x):=\sup_{a\in I}|R_m^R(a\mid x)-a|.
\]
The first accuracy assumption of the theorem gives
$q_m=O_{\Prob}(\rho_m)$. For its final claim,
Proposition~\ref{prop:risk-rate}, together with the assumptions on the
original risk-curve estimator and the fact that
$I\subset(0,R(\lambda_{\min}\mid x))$, gives
$q_m(x)=O_{\Prob}(\eta_m+h_M)$.

For each $m,n$ and $a\in\mathcal A_n$, write
\[
  \widehat r_{m,n}(a)
  :=
  \frac{1}{n}\sum_{j=1}^n
  \loss\left(\widehat f^R_{\cD_m,a}(X_j),Y_j\right).
\]
Conditional on $\cD_m$, this is an average of i.i.d. variables in $[0,B]$
with mean $r_m(a)$. Moreover, the losses are pointwise nondecreasing in $a$,
as established in the proof of Theorem~\ref{thm:marginal-validity}. This
ordering gives a uniform calibration bound without a union bound over
$\mathcal A_n$. Indeed, for $t\in[0,B]$, the sets
\[
  \left\{(x,y):
  \loss\left(\widehat f^R_{\cD_m,a}(x),y\right)>t
  \right\},
  \qquad a\in\mathcal A_n,
\]
are nested in $a$. For each fixed $t$, assign an observation the first grid
index at which its loss exceeds $t$. The displayed events are then ordinary
threshold events for this one-dimensional index. Conditional on $\cD_m$, the
Dvoretzky--Kiefer--Wolfowitz inequality therefore gives
\[
  \Prob\{D_{m,n}(t)>u\mid\cD_m\}
  \le 2\exp(-2nu^2),
\]
where
\[
  D_{m,n}(t)
  :=
  \max_{a\in\mathcal A_n}
  \left|
  \frac{1}{n}\sum_{j=1}^n
  \ind\!\left\{
  \loss\left(\widehat f^R_{\cD_m,a}(X_j),Y_j\right)>t
  \right\}
  -
  \Prob\!\left\{
  \loss\left(\widehat f^R_{\cD_m,a}(X),Y\right)>t
  \,\middle|\,\cD_m
  \right\}
  \right|.
\]
It follows that
$\E\{D_{m,n}(t)\mid\cD_m\}\le\sqrt{\pi/(2n)}$. The layer-cake
representation and Tonelli's inequality now yield
\[
  \E\left\{
  \max_{a\in\mathcal A_n}
  |\widehat r_{m,n}(a)-r_m(a)|
  \,\middle|\,\cD_m
  \right\}
  \le B\sqrt{\frac{\pi}{2n}}.
\]
The maximum changes by at most $B/n$ when one calibration observation is
replaced, so McDiarmid's inequality further gives, for every
$\delta\in(0,1)$,
\begin{equation*}
  \Prob\left\{
  \max_{a\in\mathcal A_n}
  |\widehat r_{m,n}(a)-r_m(a)|
  >
  B\left(
  \sqrt{\frac{\pi}{2n}}
  +
  \sqrt{\frac{\log(1/\delta)}{2n}}
  \right)
  \,\middle|\,\cD_m
  \right\}
  \le\delta.
\end{equation*}
Consequently,
\begin{equation}
  \max_{a\in\mathcal A_n}
  |\widehat r_{m,n}(a)-r_m(a)|
  =O_{\Prob}(n^{-1/2})
  \longrightarrow 0
  \quad\text{in probability}.
  \label{eq:uniform-calibration-convergence}
\end{equation}
The bound is uniform over every finite $\mathcal A_n$ and hence requires no
condition on the number of grid points.

The CRC constraint in \eqref{eq:ahat} is equivalently
\[
  \widehat r_{m,n}(a)
  \le
  \alpha-c_n,
  \qquad
  c_n:=\frac{B-\alpha}{n}\longrightarrow0.
\]
Let
\[
  E_{m,n}
  :=
  q_m
  +\max_{a\in\mathcal A_n}|\widehat r_{m,n}(a)-r_m(a)|
  +c_n.
\]
By assumption and \eqref{eq:uniform-calibration-convergence},
$E_{m,n}=O_{\Prob}(\rho_m+n^{-1/2})$. With probability tending to one,
$[\alpha-E_{m,n}-h_n,\alpha+E_{m,n}+h_n]\subset I$. On this event, let
$b_n^-$ be the largest grid point no greater than $\alpha-E_{m,n}$ and
$b_n^+$ the smallest grid point no less than $\alpha+E_{m,n}$. Then
\[
  \widehat r_{m,n}(b_n^-)
  \le b_n^-+q_m
  +\max_{a\in\mathcal A_n}|\widehat r_{m,n}(a)-r_m(a)|
  \le \alpha-c_n,
\]
so $b_n^-$ is feasible. Similarly,
$\widehat r_{m,n}(b_n^+)\ge\alpha+c_n>\alpha-c_n$, so $b_n^+$ is infeasible.
Monotonicity makes every grid point above $b_n^+$ infeasible, and therefore
\[
  \alpha-E_{m,n}-h_n
  \le b_n^-
  \le \widehat a
  < b_n^+
  \le \alpha+E_{m,n}+h_n.
\]
Consequently,
\[
  |\widehat a-\alpha|
  \le E_{m,n}+h_n
  =O_{\Prob}(\rho_m+n^{-1/2}+h_n).
\]

Finally, $\widehat a\in I$ with probability tending to one, and on that event
\[
  |R_m^R(\widehat a\mid x)-\alpha|
  \le q_m(x)+|\widehat a-\alpha|
  =O_{\Prob}(\rho_m+\eta_m+h_M+n^{-1/2}+h_n).
\]
Because the test pair is independent of $\cC$, the left-hand side is the
conditional-risk error displayed in the theorem. This proves both claims.
\qed

\subsection{Proof of Corollary \ref{cor:deployed-conditional}}

Let
\[
  r_*:=\operatorname*{ess\,inf}_{x\sim P_X}
  R(\lambda_{\min}\mid x).
\]
Because $\alpha<r_*$, choose an open interval $I$ containing $\alpha$ whose
closure lies in $(0,r_*)$. The deterministic argument in the proof of
Proposition~\ref{prop:risk-rate}, now using the common Lipschitz constant and
taking the essential supremum over $x$, gives
\[
  \sup_{a\in I}
  \operatorname*{ess\,sup}_{x\sim P_X}
  \left|
  \E\left\{\loss\left(\widehat f^R_{\cD_m,a}(X),Y\right)
  \,\middle|\,X=x,\cD_m\right\}-a
  \right|
  =
  O_{\Prob}(\eta_m+h_M).
\]
Integrating over $X$ verifies the marginal risk-scale condition of
Theorem~\ref{thm:deployed-conditional} with
$\rho_m=\eta_m+h_M$. Its first conclusion therefore gives
\[
  |\widehat a-\alpha|
  =
  O_{\Prob}(\eta_m+h_M+n^{-1/2}+h_n),
\]
and also places $\widehat a$ in $I$ with probability tending to one. On that
event, the preceding uniform bound and the triangle inequality give the same
rate uniformly over $x$ for the deployed conditional-risk error. Independence
of the test pair from $\cC$ identifies that error with the conditional
expectation displayed in the corollary. \qed

\subsection{Proof of Corollary \ref{cor:shift}}

Proposition~\ref{prop:approx-conditional} gives
\[
  \E_P\{\loss(\widehat f(X),Y)\mid X=x,\cD,\cC\}
  \le \widehat a+\varepsilon_M(x)
\]
for $P_X$-almost every $x$. Because $Y\mid X$ is unchanged, the same pointwise
bound holds under $Q$; because $Q_X\ll P_X$, integrating it over $Q_X$ proves
\eqref{eq:shift-estimated}. \qed

\subsection{Proof of Proposition \ref{prop:canonical}}

(a) Since $\lambda\mapsto R(\lambda\mid x)$ is non-increasing and
right-continuous and $R(\lambda_{\max}\mid x)=0$, the function
$F_x(\lambda)=1-R(\lambda\mid x)$ is non-decreasing and right-continuous on
$\Lam$, with $F_x(\lambda_{\max})=1$. Extending it to the real line by setting
$F_x(\lambda)=0$ for $\lambda<\lambda_{\min}$ and
$F_x(\lambda)=1$ for $\lambda>\lambda_{\max}$ makes it a valid cumulative
distribution function. If $R(\lambda_{\min}\mid x)<1$, the extension simply
places an atom of size $1-R(\lambda_{\min}\mid x)$ at $\lambda_{\min}$.
Thus there exists $Z_x\sim F_x$, for
which $\Prob(Z_x>\lambda)=1-F_x(\lambda)=R(\lambda\mid x)$ on $\Lam$.

(b) Directly from the definitions,
\begin{align*}
  R^{-1}(a\mid x)
  &=
  \inf\{\lambda\in\Lam:R(\lambda\mid x)\le a\}
  =
  \inf\{\lambda\in\Lam:1-R(\lambda\mid x)\ge 1-a\}\\
  &=
  \inf\{\lambda\in\Lam:F_x(\lambda)\ge 1-a\}
  =
  F_x^{-1}(1-a),
\end{align*}
where $F_x^{-1}$ denotes the generalized quantile over $\Lam$, including
the endpoint convention $F_x^{-1}(0)=\lambda_{\min}$.

(c) The first identity follows from (b) and the definition
\eqref{eq:oracle-rectified}. For the canonical problem, if $Z_x\sim F_x$ then
\[
  R_{\mathrm{can}}(\lambda\mid x)
  =
  \E\{\ind(Z_x>\lambda)\}
  =
  \Prob(Z_x>\lambda)
  =
  R(\lambda\mid x),
\]
so the canonical problem has exactly the same local risk curve as the original
problem, and therefore the same generalized inverse and the same rectification
map. \qed

\section{Worked examples: rectified predictions across CRC-style losses}
\label{app:examples}

Rectification replaces the global threshold $\lambda$ by the local threshold
$\lambda_a^R(x)=R^{-1}(a\mid x)$, so every rectified prediction
$f_a^R(x)=f_{\lambda_a^R(x)}(x)$ is a member of the original nested family.
The family determines what more protection looks like---a larger mask or label
set, a deeper list, a coarser node of a hierarchy, a wider interval, or a
smaller mask---and rectification determines how much of it each input
receives. The examples below, built from losses used in the CRC literature,
also expose the conditions this relies on: a scalar ordering of the family, a
fully protective endpoint, attainable budgets, and monotone realized losses.
Unless stated otherwise, the base model assigns scores $s_x(u)\in(0,1)$ to the
items of a finite universe $\mathcal U$ (pixels, labels, documents, or
candidate answers), and $f_\lambda(x)=\{u\in\mathcal U:s_x(u)\ge1-\lambda\}$,
$\lambda\in[0,1]$, so the least and most protective rules are
$f_0(x)=\emptyset$ and $f_1(x)=\mathcal U$.

\paragraph{Missed positives: masks, label sets, and retrieval sets.}
The false-negative proportion $\loss_{\rm FN}$ of Section~\ref{sec:crc}, with
$Y\subseteq\mathcal U$ the true positives, underlies the tumor-segmentation and
multilabel examples of \citet{angelopoulos2024conformal} and the retrieval
stage of \citet{xu2025twostage}. The rectified rule
$f_a^R(x)=\{u:s_x(u)\ge1-\lambda_a^R(x)\}$ is still a score superlevel set,
but its cutoff is input-specific. Easy images receive a high cutoff and a
tight mask, and hard images a lower cutoff and a larger mask whose boundary
follows the level sets of the segmentation score; ambiguous inputs receive
larger label sets; and queries with diffuse relevance are retrieved more deeply
than queries whose relevant documents are clearly ranked on top.

\paragraph{Modified nDCG.}
For the ranking stage of \citet[Section~4.2]{xu2025twostage}, let the response
$Z=(d_{(1)},\ldots,d_{(m)})$ list the relevant documents in decreasing order of
relevance, let $f_\lambda(x)$ be the retained documents, and define
\[
  \loss_{\rm nDCG}(f_\lambda(x),Z)
  =1-\frac{\sum_{j=1}^{m}\ind\{d_{(j)}\in f_\lambda(x)\}/\log(j+1)}
          {\sum_{j=1}^{m}1/\log(j+1)},
\]
with $\loss_{\rm nDCG}=0$ when $m=0$. The loss is graded: missing the most
relevant document costs more than missing a marginal one, so the local depth
is short when the ranker confidently places the top documents first and long
when it does not. In the full two-stage system both thresholds may be rectified,
but CRC still needs a single scalar path through the threshold pairs.

\paragraph{Hierarchical graph distance.}
\citet{angelopoulos2024conformal} score a single predicted node of a class tree
$\mathcal T$ by its graph distance $d_{\mathcal T}$ to the set
$\operatorname{anc}(y)$ of ancestors of the true leaf $y$, including $y$. For
a nested set of selected nodes $f_\lambda(x)\subseteq\mathcal T$, a natural
extension is
\[
  \loss_{\rm tree}(f_\lambda(x),y)
  =\frac1D\min_{v\in f_\lambda(x)}\min_{u\in\operatorname{anc}(y)}
  d_{\mathcal T}(v,u),
\]
where $D$ (e.g., the diameter of $\mathcal T$) normalizes the loss to $[0,1]$
and the minimum over an empty set is $D$. The loss vanishes once a selected
node lies on the ancestral path and cannot increase as the set grows.
Rectification now controls granularity: easy inputs return a small leaf-level
neighborhood, hard inputs larger subtrees or coarser internal nodes.

\paragraph{Token-level F1 for question answering.}
Let $\mathcal U$ be the candidate answers generated for question $x$ and $Y$ a
set of acceptable reference answers. Writing the best-answer loss of
\citet{angelopoulos2024conformal} for a set of references,
\[
  \loss_{\rm F1}(f_\lambda(x),Y)
  =1-\max_{z\in f_\lambda(x)}\max_{y\in Y}\operatorname{F1}_{\rm token}(z,y),
\]
with the maximum over an empty set equal to zero, rectification makes the
number of returned candidates question-specific. This loss also shows that the
fully protective endpoint of Assumption~\ref{ass:monotone} can fail:
$\loss_{\rm F1}(f_{\lambda_{\max}}(x),Y)=0$ only if some candidate attains
token F1 of one against a reference. One may instead rectify the excess loss
$\loss_{\rm F1}(f_\lambda(x),Y)-\loss_{\rm F1}(f_{\lambda_{\max}}(x),Y)$,
which has a zero endpoint but changes the controlled quantity: calibrating it
at level $\alpha$ bounds the original risk only by
$\alpha+\E\{\loss_{\rm F1}(f_{\lambda_{\max}}(X),Y)\}$. Control of the
original loss at level $\alpha$ thus requires a bound $\beta$ on this residual
risk and calibration at level $\alpha-\beta$.

\paragraph{Ordinal losses.}
For ordinal labels $\cY=\{0,\ldots,K-1\}$ and interval outputs
$f_\lambda(x)=[l_\lambda(x),u_\lambda(x)]\cap\cY$, \citet{xu2023ordinal} study
the weighted exclusion loss $h(y)\ind\{y\notin f_\lambda(x)\}$, with weights
$h(y)\in[0,B]$, and the divergence loss
$(K-1)^{-1}\min_{k\in f_\lambda(x)}|y-k|$; both are non-increasing as the
interval expands. Rectification makes the interval width local, but it only
selects a point on the pre-specified interval path and cannot change the
path's geometry: a large $h(y)$ makes omitting class $y$ costly, yet the
interval extends toward $y$ faster only if the family already does so. The
divergence loss also illustrates saturation: a singleton next to the true
label already has small loss, so $R(\lambda_{\min}\mid x)$ can fall well below
$B=1$, and larger budgets leave the rule at $f_{\lambda_{\min}}$, which is
then conservative (cf.\ Proposition~\ref{prop:oracle-conditional}).

\paragraph{False positives and shrinking masks.}
\citet{mossina2025falsepositives} control false positives by shrinking
segmentation masks, normalizing by the size of the initial mask.
For illustration we instead take the shrinking family
$f_\lambda(x)=\{u:s_x(u)>\lambda\}$, which runs from $f_0(x)=\mathcal U$ to
$f_1(x)=\emptyset$, and the false-positive fraction
$\loss_{\rm FP}(f_\lambda(x),Y)=|f_\lambda(x)\setminus Y|/(|\mathcal U\setminus
Y|\vee1)$, which is non-increasing in $\lambda$. Protection now
means removing pixels, and rectification acts as local erosion: images with
many likely false positives receive a higher threshold and a smaller mask. The
false discovery proportion $|f_\lambda(x)\setminus Y|/(|f_\lambda(x)|\vee1)$
is often more meaningful but generally not monotone along the family.
Since CRC needs monotone \emph{realized} losses, monotonizing the estimated
risk curve does not fix this; one needs a family on which the loss is
pointwise monotone, a monotone upper envelope of the loss, or a calibration
method for non-monotone losses, such as
Learn-then-Test~\citep{angelopoulos2021learn} or stability-based conformal
risk control~\citep{angelopoulos2026nonmonotonic}.

\end{document}